\documentclass[10pt,journal,compsoc]{IEEEtran}

\ifCLASSOPTIONcompsoc
  \usepackage[nocompress]{cite}
  \usepackage[caption=false,font=footnotesize,labelfont=sf,textfont=sf]{subfig}
\else
  \usepackage{cite}
  \usepackage[caption=false,font=footnotesize]{subfig}
\fi

\ifCLASSINFOpdf
  \usepackage[pdftex]{graphicx}
  \DeclareGraphicsExtensions{.pdf,.jpeg,.png}
\else
  \usepackage[dvips]{graphicx}
  \DeclareGraphicsExtensions{.eps}
\fi

\usepackage{amsmath,amssymb,amsfonts,dsfont,bm,bbm,mathrsfs,pifont}
\usepackage{algorithm,algpseudocode,listings}
\usepackage{booktabs,multirow,adjustbox,diagbox}
\usepackage{float,capt-of}
\usepackage{url}
\usepackage[pagebackref=false,breaklinks=true,colorlinks=true,citecolor=blue,bookmarks=false]{hyperref}
\usepackage[capitalize]{cleveref}  

\crefname{section}{Sec.}{Secs.}
\Crefname{section}{Section}{Sections}
\crefname{table}{Tab.}{Tabs.}
\Crefname{table}{Table}{Tables}
\crefname{figure}{Fig.}{Figs.}
\Crefname{figure}{Figure}{Figures}
\crefname{equation}{Eq.}{Eqs.}
\Crefname{equation}{Equation}{Equations}
\newcommand{\E}{\mathbb{E}}                   
\newcommand{\z}{{\rm\bf z}}                   
\newcommand{\w}{{\rm\bf w}}                   
\newcommand{\Z}{\mathcal{Z}}                  
\newcommand{\W}{\mathcal{W}}                  
\newcommand{\WP}{\mathcal{W^{+}}}             
\renewcommand{\S}{\mathcal{S}}                
\newcommand{\x}{{\rm\bf x}}                   
\newcommand{\X}{\mathcal{X}}                  
\newcommand{\n}{{\rm\bf n}}                   
\newcommand{\Loss}{\mathcal{L}}               
\newcommand{\FID}{\textbf{FID$\downarrow$}}  
\newcommand{\SWD}{\textbf{SWD$\downarrow$}}  
\newcommand{\MSE}{\textbf{MSE$\downarrow$}}  
\newcommand{\SSIM}{\textbf{SSIM$\uparrow$}}  

\newcommand{\tb}[1]{\textbf{#1}}

\usepackage{xcolor}

\definecolor{myMagenta}{rgb}{1,0,0.4}

\definecolor{myred}{RGB}{210,50,25}
\definecolor{mygreen}{RGB}{34,170,133}
\definecolor{revision_color}{rgb}{1.0, 0.5, 0.0}

\newcommand{\ReviseSecond}[1]{\textcolor{black}{#1}}

\begin{document}

\newcommand{\titlename}{In-Domain GAN Inversion for Faithful Reconstruction and Editability}
\title{\titlename}

\author{
  Jiapeng Zhu,
  Yujun Shen,
  Yinghao Xu,
  Deli Zhao, 
  Qifeng Chen, and
  Bolei Zhou, \IEEEmembership{Member, IEEE}%
  \IEEEcompsocitemizethanks{
    \IEEEcompsocthanksitem J. Zhu and Q. Chen are with
    the Department of Computer Science and Engineering, the Hong Kong University of Science and Technology, Hong Kong SAR.\protect
    \IEEEcompsocthanksitem Y. Shen and D. Zhao are with
    Ant Research, China.\protect
    \IEEEcompsocthanksitem Y. Xu is with
    the Department of Information Engineering, the Chinese University of Hong Kong, Hong Kong SAR.\protect
    \IEEEcompsocthanksitem B. Zhou is with the Computer Science Department, University of California, Los Angeles, USA. 
  }%
}

\markboth{IEEE Transactions on Pattern Analysis and Machine Intelligence}%
{Zhu \MakeLowercase{\textit{et al.}} \titlename}

\IEEEtitleabstractindextext{

\begin{abstract}
Generative Adversarial Networks (GANs) have significantly advanced image synthesis through mapping randomly sampled latent codes to high-fidelity synthesized images.
However, applying well-trained GANs to real image editing remains challenging.
A common solution is to find an approximate latent code that can adequately recover the input image to edit, which is also known as GAN inversion.
To invert a GAN model, prior works typically focus on reconstructing the target image at the pixel level, yet few studies are conducted on whether the inverted result can well support manipulation at the semantic level.
This work fills in this gap by proposing \textit{in-domain GAN inversion}, which consists of a \textit{domain-guided encoder} and a \textit{domain-regularized optimizer}, to regularize the inverted code in the native latent space of the pre-trained GAN model.
In this way, we manage to sufficiently reuse the knowledge learned by GANs for image reconstruction, facilitating a wide range of editing applications without any retraining.
We further make comprehensive analyses on the effects of the encoder structure, the starting inversion point, as well as the inversion parameter space, and observe the trade-off between the reconstruction quality and the editing property.
Such a trade-off sheds light on how a GAN model represents an image with various semantics encoded in the learned latent distribution.
Code, models, and demo are available at the project page \url{https://genforce.github.io/idinvert/}.

\end{abstract}

}

\maketitle
\IEEEdisplaynontitleabstractindextext
\IEEEpeerreviewmaketitle

\IEEEraisesectionheading{\section{Introduction}\label{sec:introduction}}

\IEEEPARstart{T}{he} rapid development of Generative Adversarial Networks (GANs) has enabled synthesizing photo-realistic images at high resolution~\cite{pggan, stylegan, stylegan2}.
The formulation of GANs~\cite{gan} makes the generator reproduce the observed data distribution through random sampling of a pre-defined latent distribution.
Existing work has confirmed that, when learning to map a latent code to a synthesis, GANs spontaneously encode versatile variation factors inside the latent space~\cite{goetschalckx2019ganalyze, gansteerability, shen2020interfacegan, shen2021closed}.
Each variation factor appears as an interpretable direction such that moving the latent code along a certain direction can result in the editing of the output image regarding a particular attribute.
However, such a manipulation capability of the latent space is hardly applicable to real images because GAN lacks inference ability.

To bridge this gap, many attempts have been made to reverse the generation process of GANs, which is widely known as GAN inversion.
Existing studies~\cite{zhu2016generative, luo2017learning, lipton2017precise, creswell2018inverting, bau2019inverting, lia, invertibility, image2stylegan} typically focus on reconstructing the pixel values of the target image, yet there is little attention on whether the inverted code supports satisfactory manipulation like the latent codes sampled from the native latent space.
Accordingly, there is no guarantee on the editing property of the inversion results, which we call the editability of the latent code, leading to limited applications in practice.

Ideally, a good GAN inversion approach should recover the input image not only from the pixel level but, more importantly, from the semantic level.
Only in this way can we reuse the knowledge (\textit{i.e.}, the variation factors encoded in the latent space) learned by GANs to manipulate the input, which is the subsequent objective of GAN inversion.
For this purpose, we propose to regularize the inverted code within the original latent space of the generator, instead of simply reconstructing the per-pixel values.
We term the resulting code as \textit{in-domain} code since it aligns with the semantic domain emerging from the pre-trained GAN model.
Concretely, we first train a \textit{domain-guided} encoder to project the image space back to the latent space such that all codes produced by the encoder are in-domain.
We then come up with a \textit{domain-regularized} optimizer, which tunes the inverted code by involving the encoder as a regularizer, to better reconstruct the pixel values without affecting the editing property.
Thanks to the in-domain inversion pipeline, which we call \textit{IDInvert}, we manage to represent a target image based on a well-trained GAN model both visually and semantically, facilitating various downstream editing tasks.

We further make an in-depth analysis of the proposed approach to explore how different design choices affect the inversion performance.
First, we find that, in comparison with the model structure and learning capacity, how the encoder is trained matters more to the property of the inverted codes.
Thus it is possible to use a lightweight encoder to accelerate the inversion process.
Second, we confirm that choosing an adequate starting point (\textit{i.e.}, the initial latent code as an inductive bias for inversion) significantly eases the training difficulty of the encoder.
Third, we study the effect of the dimension of the inversion space and observe a clear trade-off between reconstruction quality and editing property.
Concretely, enlarging the parameter space can make the image recovery more accurate, but on the other side, it weakens the alignment between the inverted code and the latent semantics learned by the pre-trained GAN model.
In other words, we can always use a well-learned generator to overfit an image, however, only a semantically meaningful representation (\textit{i.e.}, inversion) can support manipulation.

The preliminary result of this work is published in \cite{zhu2020indomain}.
In this journal extension, we add the following new content:
(i) a detailed study on the encoder architecture in \cref{subsec:encoder-architecture-explore}, which suggests that how the encoder is learned matters more than its model capacity;
(ii) an analysis on the initial inversion point for the encoder learning in \cref{subsec:add-average}, which suggests that a good starting point significantly eases the training difficulty and hence results in a better convergence;
(iii) an improved solution, based on the above analysis, to making our approach compatible with StyleGAN2~\cite{stylegan2} in \cref{subsec:add-average};
(iv) an exhaustive investigation on the trade-off between reconstruction quality and editing property in the GAN inversion task in \cref{sec:trade-off},
including (v) a discussion on whether we should retrain the generator with a larger latent space in \cref{subsec:retraining}
and (vi) an analysis on the effect of extending the inversion space with a noise space in \cref{subsec:extending-space}.
\section{Related Work}\label{subsec:related-work}

\noindent\textbf{Generative Adversarial Networks (GANs).}
GANs typically consist of a generator and a discriminator.
The generator maps latent codes to generated images, while the discriminator is used to distinguish generated images from real ones. 
These two networks are trained simultaneously in a two-player minimax game, as described in the original GAN paper \cite{gan}.
Through the efforts of communities, GANs have made tremendous progress in terms of training stability \cite{sngan,wgan,wgan_gp,whichgan} and synthesized quality \cite{pggan,biggan,stylegan,stylegan2,stylegan2ada,stylegan3}. 
The images synthesized by those works are almost indistinguishable from real ones.
Image editing in the latent space of a pre-trained GAN is one important application that benefits from this progress.

\noindent\textbf{Semantic Discovering in GANs.}
The seminal work can be traced to DCGAN \cite{dcgan}, which shows that the latent space of a GAN is correlated with the property of semantic editing. 
For instance, adding or removing the glass from an image can be fulfilled by performing a simple additive or subtractive operation in the latent space with corresponding semantic vectors.
Following that, a large number of works \cite{ramesh2019spectral, gansteerability, plumerault2020controlling, voynov2020unsupervised, shen2019interpreting, yang2019semantic, ganspace, shen2021closed, zhu2021lowrankgan} have shown that the latent spaces of GANs contain rich semantic knowledge of image attributes  and can be used to edit images via arithmetic of attribute vectors in latent spaces.
These methods can be broadly categorized into two types: \emph{supervised} and \emph{unsupervised} algorithms. 
In supervised methods, semantics can be discovered under the guidance of pre-defined attribute labels.
For example, Shen \emph{et al.} \cite{shen2019interpreting} aim at finding attribute directions corresponding to human-interpretable transformations on face images in latent space by utilizing pre-defined attribute labels on CelebA \cite{liu2015faceattributes}.
Yang \emph{et al.} \cite{yang2019semantic} uncover that hierarchical semantics emerge from layer-wise latent codes of GANs for scene synthesis by using pre-defined labels on Places \cite{zhou2017places}.
Unsupervised methods have also garnered interest from researchers as they do not require pre-defined labels.
Ramesh \emph{et al.} \cite{ramesh2019spectral} demonstrate that the leading right-singular vectors of the Jacobian matrix from Singular Value Decomposition (SVD) are the most disentangled directions, which could be used to control the generated images.
More recently, the authors of GANSpace \cite{ganspace} conducted unsupervised Principal Component Analysis (PCA) on the latent space, and then applied principal directions to achieve interpretable control over the output images.
SeFa \cite{shen2021closed} provides a closed-form solution for finding meaningful directions through eigenvalue decomposition on the weight matrix of the generator.
The recent work, LowRankGAN \cite{zhu2021lowrankgan}, demonstrates that local control over synthesized images can be achieved by shifting low-rank latent codes derived from the Jacobians of the generator in the latent space. 
However, due to the lack of inference capability in GANs, it remains challenging to apply the rich semantics encoded in the latent space to edit real images.

\definecolor{royalazure}{rgb}{0.0, 0.22, 0.66}
\begin{figure*}[t]
  \centering
  \includegraphics[width=1.0\linewidth]{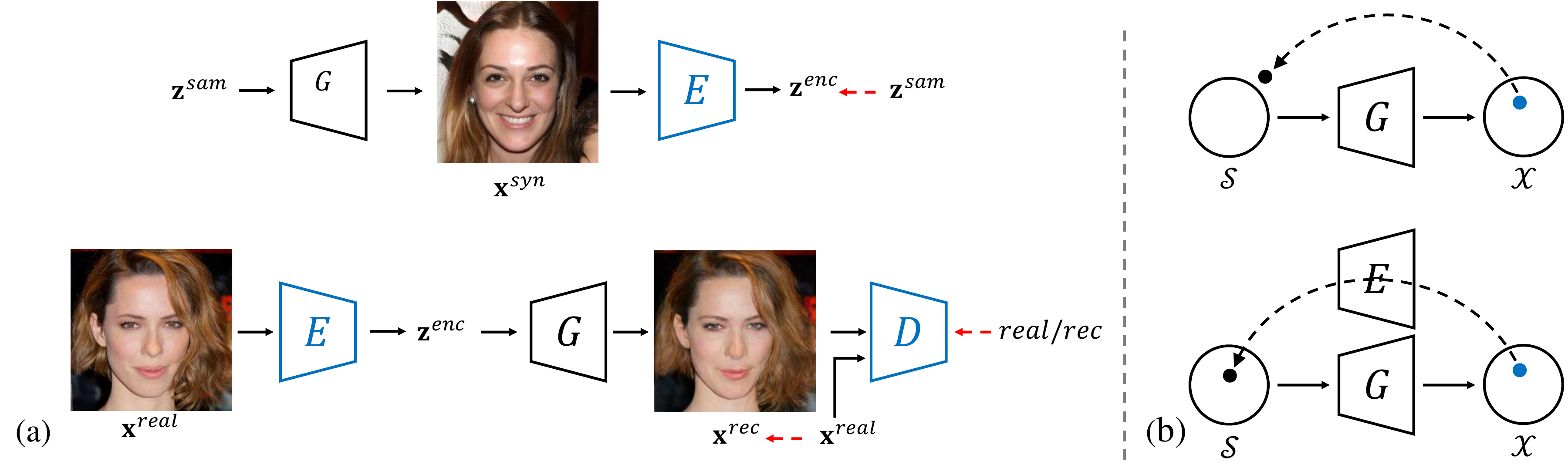}
  \vspace{-25pt}
  \caption{
    (a) The comparison between the training of conventional encoder and \emph{domain-guided} encoder for GAN inversion.
    Model blocks in \textbf{\textcolor{royalazure}{blue}} are trainable and \textbf{\textcolor{red}{red}} dashed arrows indicate the supervisions.
    Instead of being trained with synthesized data to recover the latent code, our \emph{domain-guided} encoder is trained with the objective of recovering the real images.
    The \emph{fixed} generator is involved in making sure the codes produced by the encoder lie in the native latent space of the generator and stay semantically meaningful.
    (b) The comparison between the conventional optimization and our \emph{domain-regularized} optimization.
    The well-trained \emph{domain-guided} encoder is included as a regularizer to land the latent code in the semantic domain during the optimization process.
  }
  \label{fig:framework}
  \vspace{-10pt}
\end{figure*}

\noindent\textbf{GAN Inversion.}
The goal of GAN inversion is to solve the latent code of a real image, thus enabling real image editing from the latent space \cite{zhu2016generative,bau2019semantic}.
The generator $ G $ usually maps a sampled latent code $ \z^{sam} $ to a synthesized image $ \x^{syn} $ (i.e., $ \x^{syn} = G(\z^{sam}) $), while the GAN inversion aims at finding the reverse mapping of $ G $ (i.e., $ G^{-1} $) that could map a real image $ \x^{real} $ to a latent code $ \z^{inv} $ (i.e., $ \z^{inv} = G^{-1}( \x^{real} ) $).
Existing inversion approaches typically fall into three categories.
The first one is encoder-based \cite{ali, bigan, donahue2019bigbigan,perarnau2016invertible, zhu2016generative, bau2019semantic, lia, richardson2021encoding, xu2021generative, alaluf2021restyle, tov2021designing, bdinvert, wang2021HFGI, dinh2021hyperinverter, bai2022high}, which learns an encoder $ E $ to approximate the inversion of a generator (i.e., $ E = G^{-1} $).
Among them, \cite{ali, bigan, donahue2019bigbigan} learns $ G $ and $ E $ simultaneously. 
These methods primarily focus on representation learning, which emphasizes the high-level semantics of images over reconstruction precision.
Therefore, the inverted codes obtained from these methods may not produce a faithful reconstruction.
Some methods, such as \cite{perarnau2016invertible, zhu2016generative}, first synthesize a collection of images with randomly sampled latent codes and then use the images and codes as inputs and supervisions, respectively, to train the encoder.
However, a drawback of these methods is that the encoder never sees real images during the entire training process, which can result in relatively poor reconstruction results on real images.
The pSp algorithm \cite{richardson2021encoding} uses a feature pyramid encoder architecture as well as an identity loss to fulfill various image-to-image translation tasks.
Due to identity loss, this method is limited to face data.
In recent years, \cite{bdinvert, wang2021HFGI, dinh2021hyperinverter, bai2022high} involve additional spaces (e.g., feature space, parameter space, padding space) to recover a given image accurately.
The second one is optimization-based, which deals with a single instance at one time by directly optimizing the latent code to minimize the pixel-wise reconstruction loss \cite{lipton2017precise,creswell2018inverting,image2stylegan,image2stylegan++,gu2020image,stylegan2, pan2020exploiting, huh2020ganprojection, roich2021pivotal}.
Some use a random latent code as the initialization for optimization~\cite{lipton2017precise,creswell2018inverting}, which can make it difficult to obtain a good reconstruction result due to the optimization process being highly sensitive to the initial value.
Gu \emph{et al.} \cite{gu2020image} addresses this sensitivity by employing multiple latent codes to recover a target image, which leads to satisfying reconstruction results. 
Abdal \emph{et al.} \cite{image2stylegan} uses the mean value of the $ W $ distribution, which is generated by the generator during training, as the initialization substantially improves reconstruction results.
Karras \emph{et al.} \cite{stylegan2} and Abdal \emph{et al.} \cite{image2stylegan++} show that utilizing layer-wise noises when inverting the StyleGAN \cite{stylegan} models could promote reconstruction performance.
Pan \emph{et al.} \cite{pan2020exploiting} and Huh \emph{et al.} \cite{huh2020ganprojection} explore the optimization-based method on BigGAN \cite{bigan}. 
The last approach combines these two ideas into a two-step method~\cite{zhu2016generative,bau2019inverting, roich2021pivotal}, which uses a trained encoder to generate a starting point for the optimization process.
However, their two-step method primarily focuses on achieving good reconstruction results but overlooks the importance of maintaining the resulting latent code within the native latent space. (\textit{i.e.}, without any constraints during optimization, the final latent code can deviate significantly from the starting point, even if it is a good starting point).
Hence, to ensure that the latent code remains within the native latent space, we add an encoder as a regularizer during optimization. 
One crucial issue is that existing inversion methods merely focus on reconstructing the target image at the pixel level without considering semantic information in inverted codes.
This work fills in this gap by studying the property of inverted codes, e.g., whether the inverted codes could support semantic editing.
Moreover, we find a trade-off between reconstruction quality and editing property by analyzing the inversion parameter space, revealing that it is not easy to improve both simultaneously.

\noindent\textbf{Image-to-Image Translation.}
Image-to-Image translation task aims at learning the mappings between different domains.
A seminal work can be traced to \cite{pix2pix2017}, which uses a conditional GAN \cite{cgan} to perform various image-to-image translation tasks.
After that, a larger number of works have been proposed to deal with this task, varying from the supervised method \cite{wang2018pix2pixHD, park2019SPADE} to unsupervised method \cite{CycleGAN2017, huang2018munit, DRIT}, from single-modal output \cite{guidedI2I, DiscoGAN} to multi-modal outputs \cite{choi2018stargan, yu2019multi, choi2020starganv2}.
However, the image-to-image translation task together learns the encoder, decoder, and desired manipulation.
Rather, GAN inversion learns an encoder to a pre-trained generator, and all the manipulations depend on the pre-trained generator.

\section{In-Domain GAN Inversion}\label{sec:method}
As discussed above, besides recovering the input image by pixel values, we also care about whether the inverted code is semantically meaningful when inverting a GAN model.
Here, the semantics refer to the emergent knowledge that GAN has learned from the observed data \cite{goetschalckx2019ganalyze,gansteerability,shen2019interpreting,yang2019semantic}.
For this purpose, we propose to first train a \emph{domain-guided} encoder and then use this encoder as a regularizer for the further \emph{domain-regularized} optimization, as shown in \cref{fig:framework}.

\noindent\textbf{Problem Statement.}
Before going into details, we briefly introduce the problem setting with some basic notations.
A GAN model typically consists of a generator $G(\cdot): \Z\rightarrow\X$ to synthesize high-quality images and a discriminator $D(\cdot)$ to distinguish real from synthesized data.
GAN inversion studies the reverse mapping of $G(\cdot)$, which is to find the best latent code $\z^{inv}$ to recover a given real image $\x^{real}$.
We denote the semantic space learned by GANs as $\S$.
We would like $\z^{inv}$ to also align with the prior knowledge $\S$ in the pre-trained GAN model.

\noindent\textbf{Choice of Latent Space.}
Typically, GANs sample latent codes $\z$ from a pre-defined distributed space $\Z$, such as normal distribution.
The recent StyleGAN model \cite{stylegan} proposes to first map the initial latent space $\Z$ to a second latent space $\W$ with Multi-Layer Perceptron (MLP), and then feed the codes $\w\in\W$ to the generator for image synthesis.
Such additional mapping has already been proven to learn more disentangled semantics \cite{stylegan,shen2019interpreting}.
As a result, the disentangled space $\W$ is widely used for the GAN inversion task \cite{image2stylegan,lia,image2stylegan++,stylegan2}.
Similarly, we also choose $\W$ space as the inversion space for three reasons:
(i) We focus on the semantic (\emph{i.e.,} \emph{in-domain}) property of the inverted codes, making $\W$ space more appropriate for analysis.
(ii) Inverting to $\W$ space achieves better performance than $\Z$ space \cite{lia}.
(iii) It is easy to introduce the $\W$ space to any GAN model by simply learning an extra MLP ahead of the generator. Hence, it will not harm the generalization ability of our approach.
In this work, we conduct all experiments on the $\W$ space, but our approach can be performed on the $\Z$ space as well.
For simplicity, we use $\z$ to denote the latent code in the following sections.

\subsection{Domain-Guided Encoder}\label{subsec:domain-guided-encoder}
Training an encoder is commonly used for GAN inversion problem \cite{perarnau2016invertible,zhu2016generative,bau2019inverting} considering its fast inference speed.
However, existing methods simply learn a deterministic model with no regard to whether the codes produced by the encoder align with the semantic knowledge learned by $G(\cdot)$.
As shown on the top of \cref{fig:framework}(a), a collection of latent codes $\z^{sam}$ are randomly sampled and fed into $G(\cdot)$ to get the corresponding synthesis $\x^{syn}$.
Then, the encoder $E(\cdot)$ takes $\x^{syn}$ and $\z^{sam}$ as inputs and supervisions respectively and is trained with
\begin{align}
  \min_{\Theta_E}\Loss_E = ||\z^{sam} - E(G(\z^{sam}))||_2, \label{eq:conventional-encoder}
\end{align}
where $||\cdot||_2$ denotes the $l_2$ distance and $\Theta_E$ represents the parameters of the encoder $E(\cdot)$.
We argue that the supervision by only reconstructing $\z^{sam}$ is not powerful enough to train an accurate encoder.
Also, the generator is actually omitted and cannot provide its domain knowledge to guide the training of encoder since the gradients from $G(\cdot)$ are not taken into account at all.

To solve these problems, we propose to train a \emph{domain-guided} encoder, which is illustrated in the bottom row of \cref{fig:framework}(a).
There are three main \textbf{differences} compared to the conventional encoder:
(i) The output of the encoder is fed into the generator to reconstruct the input image such that the objective function comes from the image space instead of latent space. This involves semantic knowledge from the generator in training and provides more informative and accurate supervision. The output code is therefore guaranteed to align with the semantic domain of the generator.
(ii) Instead of being trained with synthesized images, the \emph{domain-guided} encoder is trained with real images, making our encoder more applicable to real applications.
(iii) To make sure the reconstructed image is realistic enough, we employ the discriminator to compete with the encoder. In this way, we can acquire as much information as possible from the GAN model (\emph{i.e.}, both two components of GAN are used). The adversarial training manner also pushes the output code to better fit the semantic knowledge of the generator.
We also introduce perceptual loss \cite{johnson2016perceptual} using the feature extracted by VGG \cite{vgg}.
Hence, the training process can be formulated as
\begin{align}
  &\begin{aligned}
    \min_{\Theta_E}\Loss_E =\ &||\x^{real} - G(E(\x^{real}))||_2 \\
                              &+\lambda_{vgg} ||F(\x^{real}) - F(G(E(\x^{real})))||_2 \\
                              &-\lambda_{adv} \underset{\x^{real}\sim P_{data}}\E[D(G(E(\x^{real})))], \label{eq:encoder}
  \end{aligned} \\
  &\begin{aligned}
    \min_{\Theta_D}\Loss_D =\ &\underset{\x^{real}\sim P_{data}}\E[D(G(E(\x^{real})))] \\
                              &-\underset{{\x^{real}\sim P_{data}}}\E[D(\x^{real})] \\
                              &+\frac{\gamma}{2}\underset{{\x^{real}\sim P_{data}} }{\E}[||\nabla_{{\x}}D(\x^{real})||_2^2],
  \end{aligned} \label{eq:discriminator}
\end{align}
where $P_{data}$ denotes the distribution of real data and $\gamma$ is the hyper-parameter for the gradient regularization.
$\lambda_{vgg}$ and $\lambda_{adv}$ are the perceptual and discriminator loss weights.
$F(\cdot)$ denotes the VGG feature extraction model.
It is worth noting that this type of reconstruction loss is commonly utilized in other tasks, such as in image-to-image translation~\cite{pix2pix2017}, image super-resolution~\cite{wang2018esrgan, ledig2017photo}, \textit{etc}, owing to its ability to reconstruct the input images effectively.

\subsection{Domain-Regularized Optimization}\label{subsec:domain-regularized-optimization}
Unlike the generation process of GANs which learns a mapping at the distribution level, \emph{i.e.} from latent distribution to real image distribution, GAN inversion is more like an instance-level task, which is to best reconstruct a given individual image.
From this point of view, it is hard to learn a perfect reverse mapping with an encoder alone due to its limited representation capability.
Therefore, even though the inverted code from the proposed \emph{domain-guided} encoder can well reconstruct the input image based on the pre-trained generator and ensure the code itself is semantically meaningful, we still need to refine the code to make it better fit the target individual image at the pixel values.

Previous methods \cite{creswell2018inverting,invertibility,image2stylegan} propose to gradient descent algorithm to optimize the code.
The top row of \cref{fig:framework}(b) illustrates the optimization process where the latent code is optimized ``freely'' based on the generator only.
It may very likely produce an out-of-domain inversion since there are no constraints on the latent code at all.
Relying on our \emph{domain-guided} encoder, we design a \emph{domain-regularized} optimization with two improvements, as shown at the bottom of \cref{fig:framework}(b):
(i) We use the output of the \emph{domain-guided} encoder as an ideal starting point which avoids the code from getting stuck at a local minimum and also significantly shortens the optimization process.
(ii) We include the \emph{domain-guided} encoder as a regularizer to preserve the latent code within the semantic domain of the generator.
To summarize, the objective function for optimization is
\begin{align}
  \begin{aligned}
    \z^{inv} = \arg\min_{\z}\ ||\x - G(\z)||_2\ &+ \lambda_{vgg}||F(\x) - F(G(\z))||_2\ \\
                                                &+ \lambda_{dom}||\z - E(G(\z))||_2,
  \end{aligned} \label{eq:optimization}
\end{align}
where $\x$ is the target image to invert, $\lambda_{vgg}$ and $\lambda_{dom}$ are the loss weights corresponding to the perceptual loss and the encoder regularize, respectively.

\setlength{\tabcolsep}{6.5pt}
\begin{table*}[t]
  \caption{
    Quantitative comparison between different inversion methods.
    For each model, we invert 5000 images for evaluation.
    $\downarrow$ means lower number is better.
  }
  \vspace{-10pt}
  \label{tab:reconstruction_quat}
  \scriptsize\centering
  \begin{tabular}{l|cccc|cccc|cccc}
                \toprule
                & \multicolumn{4}{c|}{Face}
                & \multicolumn{4}{c|}{Tower}
                & \multicolumn{4}{c}{Bedroom}    \\
                \midrule
   \diagbox[width=16em]{Methods}{Metrics}
                &    \FID    &    \SWD    &    \SSIM   &   \MSE 
                &    \FID    &    \SWD    &    \SSIM   &   \MSE
                &    \FID    &    \SWD    &    \SSIM   &   \MSE    \\ 
                \midrule
    Conventional Encoder \cite{zhu2016generative}
                &      50.11 &      83.49 &    0.282   &   0.464 
                &      39.38 &      74.94 &    0.293   &   0.525
                &      61.60 &      84.38 &    0.196   &   0.681     \\
    Image2StyleGAN \cite{image2stylegan}
                &      25.35 &      27.31 & \tb{0.702} & \tb{0.026}  
                &      39.54 &      58.12 & \tb{0.594} & \tb{0.061}      
                &      43.17 &      61.40 & \tb{0.684} & \tb{0.036}  \\ 
                \midrule
    Domain-Guided Encoder (Ours)
                &      16.43 & \tb{13.02} &    0.522   &    0.071 
                & \tb{12.41} &      19.50 &    0.500   &    0.093
                &  \tb{9.61} &      21.93 &    0.521   &    0.079    \\
    In-Domain Inversion (Ours)
                & \tb{15.07} &     14.92  &    0.568   &   0.053 
                &      12.47 & \tb{15.17} &    0.530   &   0.087
                &      11.07 & \tb{20.90} &    0.560   &   0.066    \\ 
                \bottomrule
  \end{tabular}
  \vspace{-10pt}
\end{table*}

\section{Evaluation Tasks}\label{sec:evaluation-metrics-and-applications}
In this section, we introduce the tasks used to evaluate our proposed method, including its inversion quality and several image editing applications.
But, we give the implementation details for those tasks being assessed first.

\subsection{Implementation Details}\label{subsec:implementation-details}
We conduct experiments on the FFHQ dataset \cite{stylegan}, and three sub-categories on LSUN dataset \cite{yu2015lsun}, i.e., tower (outdoor scene), church(outdoor scene), and bedroom (indoor scene), to evaluate our proposed method including the \emph{domain-guided encoder} and \emph{domain-regularized optimization}.
For FFHQ, which contains 70,000 high-quality face images, we take the first 65,000 faces as the training set and the remaining 5,000 faces as the reconstruction test according to the exact order of the dataset.
And for two subsets in LSUN, the 0.1 million images are selected by random sample from the first 0.6 million images in each dataset and 5,000 images are selected as the test data from the remaining dataset, respectively. 
The GANs to invert are pre-trained following StyleGAN \cite{stylegan}.\footnote{Different from StyleGAN, we use different latent codes for different layers in model training. More discussions can be found in \cref{subsec:retraining}.}
When training the domain-guided encoder, the generator is \emph{fixed}, and we only update the encoder and discriminator according to \cref{eq:encoder} and \cref{eq:discriminator}, and the loss weights are set as $\lambda_{vgg}=5e^{-5}$, $\lambda_{adv}=0.1$, and $\gamma=10$ in these two equations.
As for the perceptual loss in \cref{eq:encoder}, we take $\mathtt{conv4\_3}$ as the VGG \cite{vgg} output.
We set $\lambda_{dom}=2$ in \cref{eq:optimization} for the \emph{domain-regularized} optimization, an ablation study on this parameter can be found in \cref{subsec:ablation-study}.
For the encoder architecture, we will give a detailed analysis in \cref{sec:archtecture-exploring}.
And for quantitative evaluation metrics, we use Fr\'{e}chet Inception Distance (FID), Sliced Wasserstein Distance (SWD), Mean-Squared Error (MSE),  and Structural Similarity Index Measure (SSIM).
These metrics are commonly used to measure the quality of the images in GANs \cite{pggan, bigan, stylegan, stylegan2}.

\begin{figure}[t]
  \centering
  \includegraphics[width=1.0\linewidth]{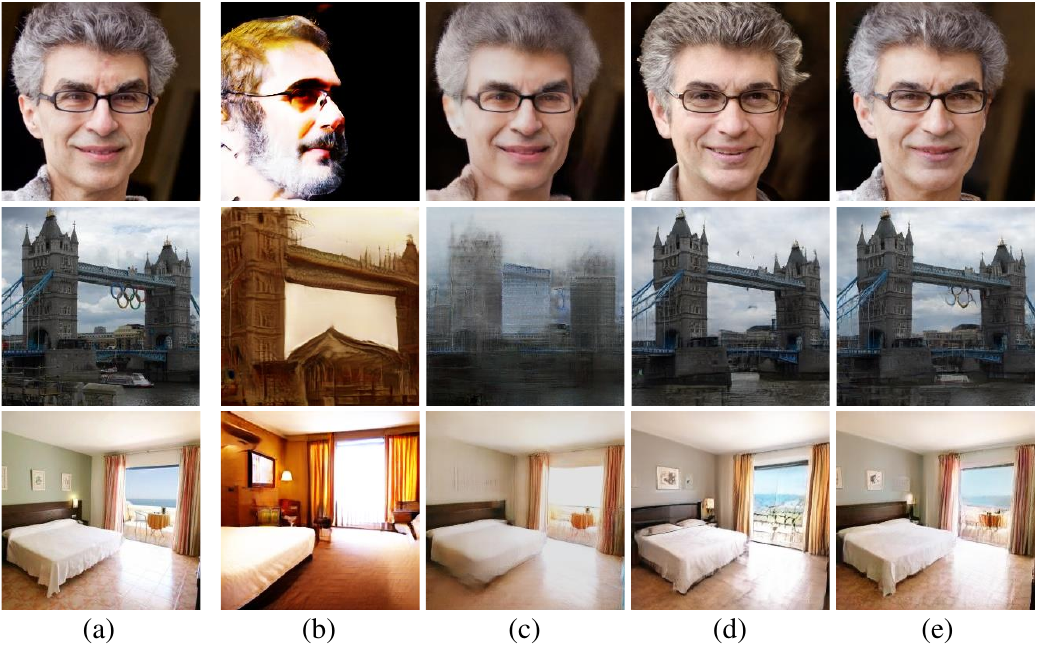}
  \vspace{-20pt}
  \caption{
    Qualitative comparison on image reconstruction with different GAN inversion methods.
    (a) Input image.
    (b) Conventional encoder \cite{zhu2016generative}.
    (c) Image2StyleGAN \cite{image2stylegan}.
    (d) Our proposed \emph{domain-guided} encoder.
    (e) Our proposed \emph{in-domain} inversion.
  }
  \label{fig:inversion}
  \vspace{-10pt}
\end{figure}

\subsection{Inversion Quality}\label{subsec:inversion-quality}
Recall we argue that a good GAN inversion method should not only precisely recover the given images but also should keep the inverted codes semantically meaningful for the downstream tasks.
Therefore, the evaluation for this part can be divided into two parts.
One is the quality of the reconstructed images, such as their precision and realism in contrast to the input.
The other one is the property of the inverted codes, such as the alignment of the semantics with the boundaries \cite{shen2019interpreting}.
Namely, the more alignment with the boundaries, the more satisfying editing results we could obtain.

\noindent\textbf{Quality of the Reconstructed Images.}
The primary goal for GAN inversion is could faithfully reconstruct the input image.
Therefore, in this section, we evaluate the quality of the reconstructed images by our method, as well as some comparisons with the existing baselines, including traditional encoder \cite{zhu2016generative}, and MSE-based optimization \cite{image2stylegan}.
\cref{fig:inversion} shows the reconstruction results using different methods on Face, Tower, and Bedroom.
Comparison between \cref{fig:inversion}(b) and \cref{fig:inversion}(d) shows the superiority of our \emph{domain-guided} encoder in learning a better mapping from the image space to the latent space.
\cref{fig:inversion}(c) shows the blurriness when inverting the given images in contrast to the \cref{fig:inversion}(d).
And \cref{fig:inversion}(e) is our whole algorithm, which demonstrates we could further improve the reconstruction result when performing optimization as introduced in \cref{subsec:domain-regularized-optimization}.
\cref{tab:reconstruction_quat} gives the quantitative comparison results on the test data.
From the table, we could see that the Image2StyleGAN \cite{image2stylegan} surpasses our method in terms of the SSIM and MSE.
However, it just focused on reconstructing the pixels of the input image, and when it comes to fidelity and realism of the reconstructed images, our method surpasses it by a large margin (e.g., see the FID and SWD metrics).
Moreover, Image2StyleGAN takes too much time to invert an image.
For instance, it will take 290 seconds, while our method only takes 8 seconds when inverting an image on a 2080 Ti GPU.

\begin{figure}[t]
  \centering
  \includegraphics[width=1.0\linewidth]{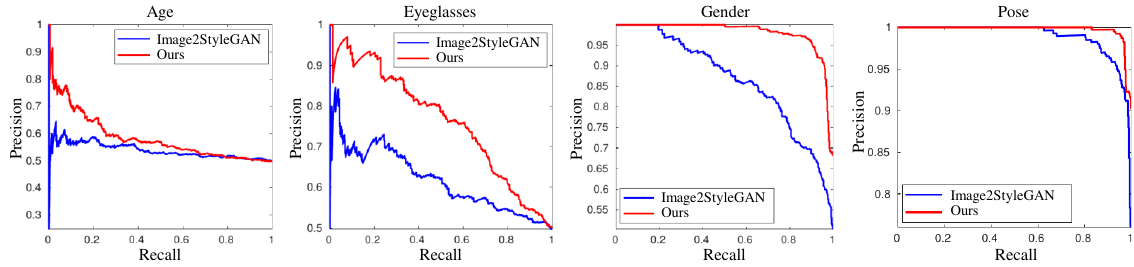}
  \vspace{-20pt}
  \caption{
    Precision-recall curves by directly using the inverted codes for facial attribute classification.
    Our \emph{in-domain} inversion shows much better performance than Image2StyleGAN \cite{image2stylegan}, suggesting a stronger semantic preservation.
  }
  \label{fig:domain}
  \vspace{-10pt}
\end{figure}

\begin{figure*}[!ht]
  \centering
  \includegraphics[width=1.0\linewidth]{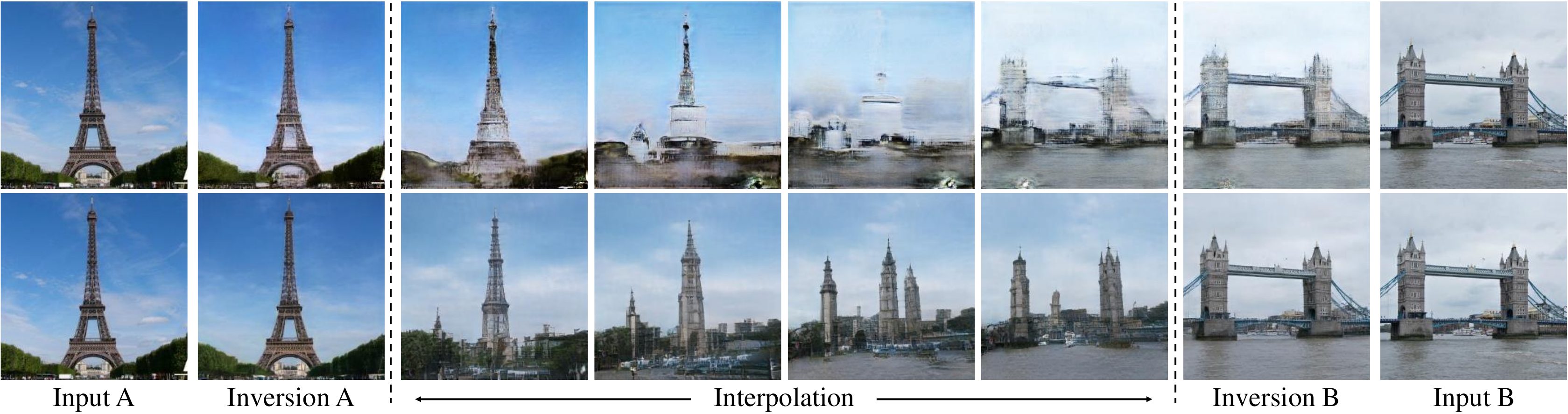}
  \vspace{-22pt}
  \caption{
    Qualitative comparison on image interpolation between Image2StyleGAN \cite{image2stylegan} (first row) and our \emph{in-domain} inversion (second row).
  }
  \label{fig:interpolation}
  \vspace{-12pt}
\end{figure*}

\noindent\textbf{Property of the Inverted Codes.}
As discussed above, our method can produce satisfying reconstruction results.
In this part, we further evaluate how the inverted codes can semantically represent the target images.
As pointed out by prior work \cite{stylegan,shen2019interpreting}, the latent space of GANs is linearly separable in terms of semantics.
In particular, for a binary attribute (\emph{e.g.}, male \emph{v.s.} female), it is possible to find a latent hyperplane such that all points from the same side correspond to the same attribute.
We use this property to evaluate the alignment between the inverted codes and the latent semantics.
We use off-the-shelf attribute classifiers to predict age (young \emph{v.s.} old), gender (female \emph{v.s.} male), eyeglasses (absence \emph{v.s.} presence), and pose (left \emph{v.s.} right) on the face test dataset.
These predictions are considered as ground-truth.
Then, we use the state-of-the-art GAN inversion method, Image2StyleGAN \cite{image2stylegan}, and our proposed \emph{in-domain} GAN inversion to invert these images back to the latent space of a \emph{fixed} StyleGAN model trained on FFHQ dataset \cite{stylegan}.
InterFaceGAN \cite{shen2019interpreting} is used to search the semantic boundaries for the aforementioned attributes in the latent space.
Then, we use these boundaries as well as the inverted codes to evaluate the attribute classification performance.
\cref{fig:domain} shows the precision-recall curves on each semantic.
We can easily tell that the codes inverted by our method are more semantically meaningful.
This quantitatively demonstrates the effectiveness of our proposed \emph{in-domain} inversion for preserving the semantics property of the inverted code.

\begin{figure}[t]
  \centering
  \includegraphics[width=0.99\linewidth]{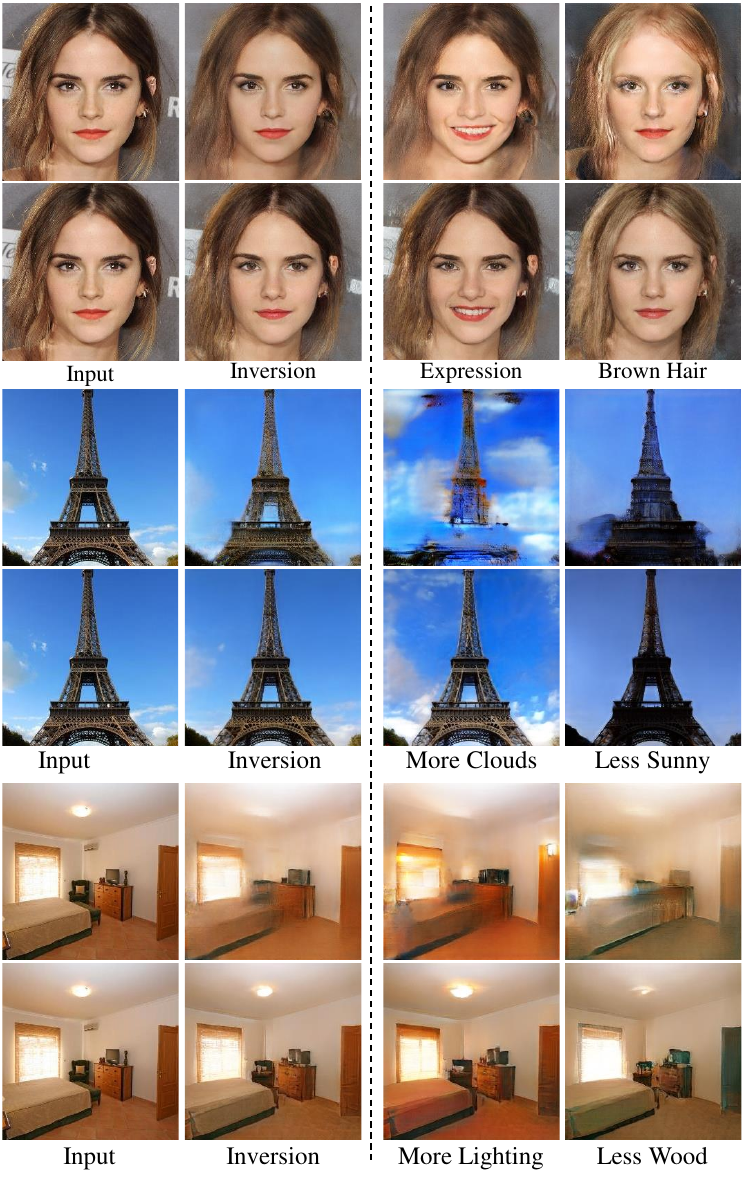}
  \vspace{-12pt}
  \caption{
    Comparison results on manipulating different datasets between Image2StyleGAN \cite{image2stylegan} (odd rows) and our \emph{in-domain} GAN inversion (even row).
  }
  \label{fig:manipulation}
  \vspace{-10pt}
\end{figure}

\subsection{Real Image Editing}\label{subsec:editing}
In this section, we evaluate our \emph{in-domain} GAN inversion approach on real image editing tasks, including image interpolation and semantic image manipulation.
We also came up with a novel image editing task, called \emph{semantic image diffusion}, to see how our approach is able to adapt the content from one image into another and keep the results semantically meaningful and seamlessly compatible.

\noindent\textbf{Image Interpolation.}
Image interpolation aims at semantically interpolating two images, which is suitable for investigating the semantics contained in the inverted codes.
In other words, for a good inversion, the semantics should vary continuously when interpolating two inverted codes.
\cref{fig:interpolation} shows the comparison results on the image interpolation task between Image2StyleGAN \cite{image2stylegan} and our \emph{in-domain} inversion.
We do experiments on the face, tower, and bedroom datasets (For the results on the face and bedroom, please refer to \cref{fig:interpolation-ffhq-bedroom}.) to analyze the semantic property more comprehensively.
For tower images, the interpolation results from Image2StyleGAN show unsatisfying artifacts.
Meanwhile, some interpolations made by Image2StyleGAN are not semantically meaningful (e.g., interpolated images are no longer a tower anymore).
On the contrary, our inverted codes lead to more satisfying interpolation.
One noticeable thing is that during interpolating two towers with different types (\emph{e.g.}, one with one spire and the other with multiple spires), the interpolated images using our approach are still high-quality towers.
This demonstrates the \emph{in-domain} property of our algorithm.
Quantitative evaluation is given in \cref{tab:editing-comparison}, which shows that our method vastly surpasses the Image2StyleGAN as well.

\setlength{\tabcolsep}{5pt}
\begin{table*}[t]
  \caption{
    Quantitative comparison on image interpolation and manipulation between Image2StyleGAN \cite{image2stylegan} and our \emph{in-domain} inversion on different datasets.
    $\downarrow$ means the lower number is better.
  }
  \vspace{-10pt}
  \label{tab:editing-comparison}
  \scriptsize\centering
  \begin{tabular}{l|cc|cc|cc|cc|cc|cc}
             \toprule
             &   \multicolumn{6}{c|}{Interpolation}                   
             &   \multicolumn{6}{c}{Manipulation}    \\
             \midrule
 Datasets    &   \multicolumn{2}{c|}{Face}
             &   \multicolumn{2}{c|}{Tower}
             &   \multicolumn{2}{c|}{Bedroom}
             &   \multicolumn{2}{c|}{Face}
             &   \multicolumn{2}{c|}{Tower}
             &   \multicolumn{2}{c}{Bedroom}    \\
             \midrule
 Metrics     &     \FID   &    \SWD    &    \FID    &    \SWD    &    \FID    &   \SWD 
             &     \FID   &    \SWD    &    \FID    &    \SWD    &    \FID    &   \SWD    \\
             \midrule
 Image2StyleGAN \cite{image2stylegan}
             &     77.54  &      30.98 &     82.37  &    68.41   &   50.46    &   60.60 
             &     31.56  &      27.46 &     76.35  &    52.54   &   67.10    &   60.58  \\
 IDInvert (Ours)
             & \tb{46.56} & \tb{25.49} & \tb{33.02} & \tb{29.74} & \tb{19.63} & \tb{21.40} 
             & \tb{17.36} & \tb{17.23} & \tb{19.87} & \tb{31.84} & \tb{15.36} & \tb{28.42} \\    
             \bottomrule
  \end{tabular}
  \vspace{-10pt}
\end{table*}

\begin{figure*}[t]
  \centering
  \includegraphics[width=0.96\linewidth]{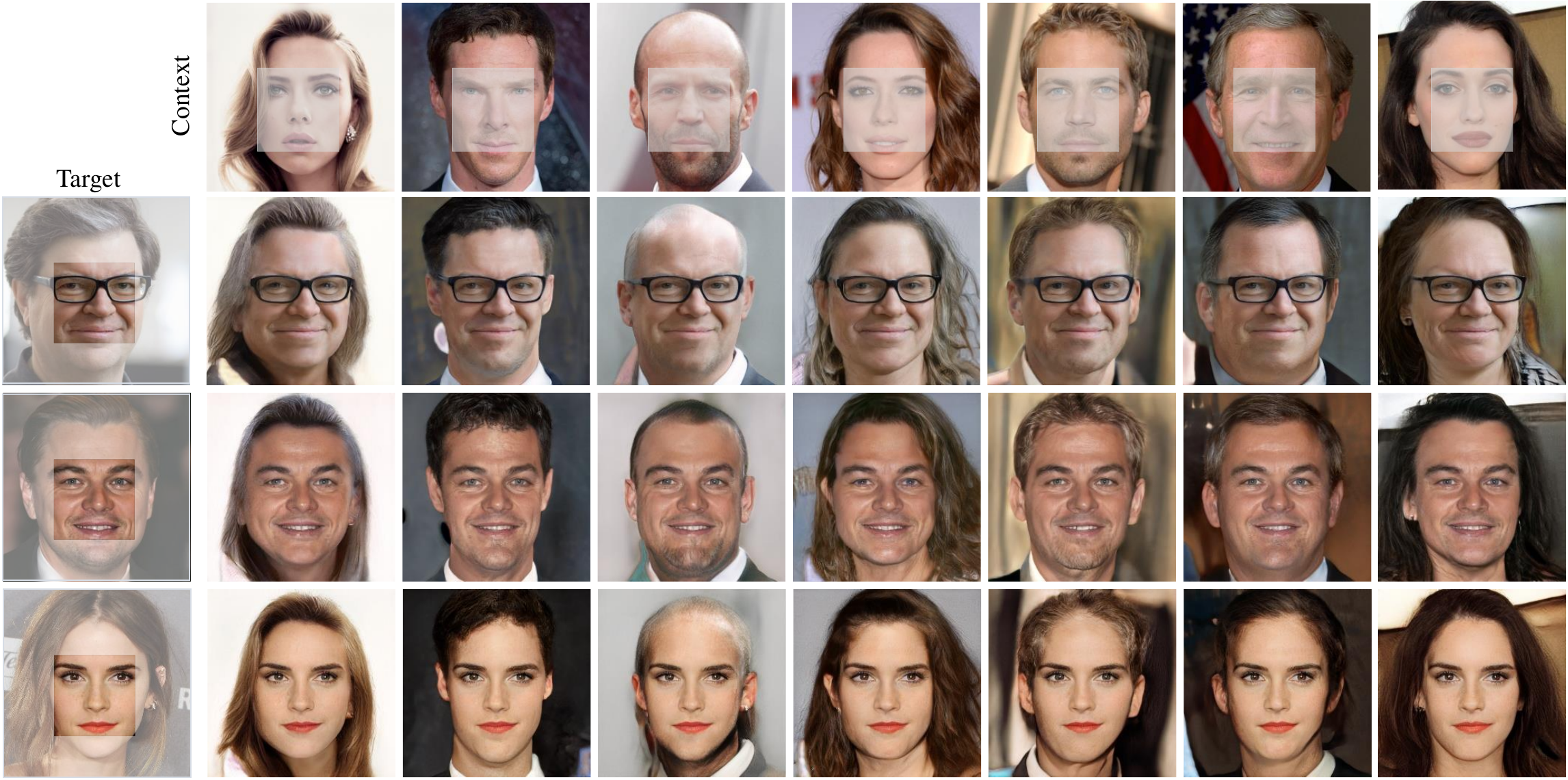}
  \vspace{-10pt}
  \caption{
    Semantic diffusion result using our \emph{in-domain} GAN inversion.
    Target images (first column) are seamlessly diffused into context images (first row) while the identity remains the same as the target. A Google Colab demo is available \href{https://colab.research.google.com/github/genforce/idinvert_pytorch/blob/master/docs/Idinvert.ipynb}{here}. 
  }
  \label{fig:face-diffusion}
  \vspace{-10pt}
\end{figure*}

\noindent\textbf{Semantic Manipulation.}
Image manipulation is another way to examine whether the embedded latent codes align with the semantic knowledge learned by GANs.
As pointed out by prior work \cite{shen2019interpreting,yang2019semantic}, GANs can learn rich semantics in the latent space, enabling image manipulation by linearly transforming the latent representation.
This can be formulated as
\begin{align}
  \x^{edit} = G(\z^{inv} + \alpha\n), \label{eq:manipulation}
\end{align}
where $\n$ is the normal direction corresponding to a particular semantic in the latent space and $\alpha$ is the step for manipulation.
In other words, if a latent code is moved in this direction, the semantics contained in the output image should vary accordingly.
We follow \cite{shen2019interpreting} to search the semantic direction $\n$.

\cref{fig:manipulation} shows the comparison results of manipulating faces, towers, and bedrooms using Images2StyleGAN \cite{image2stylegan} and our \emph{in-domain} GAN inversion (For the results on the church, please refer to \cref{fig:church-inversion}).
We can see that our method shows more satisfying manipulation results than Image2StyleGAN.
Taking face manipulation in \cref{fig:manipulation} as an example, the hair of the actress becomes blurred after adding the smile, and the identity changes a lot when editing the hair color with the codes from Image2StyleGAN.
That is because it only focuses on the reconstruction of the per-pixel values yet omits the semantic information contained in the inverted codes.
By contrast, our \emph{in-domain} inversion can preserve most other details when editing a particular facial attribute.
As for tower manipulation, we observe from \cref{fig:manipulation} that our \emph{in-domain} approach surpasses MSE-based optimization by both decreasing and increasing the semantic level.
For example, when adding clouds in the sky, Image2StyleGAN will blur the tower together with the sky, since it only recovers the image at the pixel level without considering the semantic meaning of the recovered objects.
Therefore, the cloud is added to the entire image regardless of whether a particular region belongs to the sky or tower.
Rather, our algorithm barely affects the tower itself when editing clouds, suggesting that our \emph{in-domain} inversion can produce semantically informative latent codes for image reconstruction.
And for the bedroom manipulation, the results attained by Image2StyleGAN show artifacts and blurs when adding or removing a specific semantic.
We also include the quantitative evaluation on the manipulation task in \cref{tab:editing-comparison}.
We can tell that our \emph{in-domain} inversion outperforms Image2StyleGAN from all evaluation metrics.

\noindent\textbf{Semantic Diffusion.} 
Semantic diffusion aims at diffusing a particular part (usually the most representative part) of the target image into the context of another image.
We would like the fused result to keep the characteristics of the target image (\emph{e.g.}, identity of face) and adapt the context information at the same time.
To be specific, given a target-context image pair, we first crop the wanted part from the target image and then paste it onto the context image.
Then, we use our \emph{domain-guided} encoder to infer the latent code for the stitched image.
Due to the domain-alignment property of our encoder, the reconstruction from the code can already capture the semantics from both the target patch and its surroundings and further smooth the contents.
With this code as an initialization, we finally perform masked optimization by only using the target foreground region to compute the reconstruction loss.
In this way, we are able to not only diffuse the target image to any other context but also keep the original style of the context image.
\cref{fig:face-diffusion} shows some examples where we successfully diffuse various target faces into diverse contexts using our \emph{in-domain} GAN inversion approach.
We can see that the results well preserve the identity of the target face and reasonably integrate into different surroundings.
This is different from style mixing since the center region of our resulting image is kept the same as that of the target image.
More detailed analysis of the semantic diffusion operation can be found in the \cref{sec:supp-semantic-diffusion}.

\setlength{\tabcolsep}{6pt}
\begin{table*}[t]
  \setlength{\tabcolsep}{9.0pt}
  \caption{
         Quantitative results of various encoder architecture on FFHQ, and Church datasets. For the \tb{Speed}, the unit is second.
  }
  \label{tab:encoder-arch}
  \vspace{-8pt}
  \scriptsize\centering
  \begin{tabular}{l|cccc|cccc|c}
    \toprule
                       Datasets   &
    \multicolumn{4}{c|}{Face}     &
    \multicolumn{4}{c}{Church}       \\
    \midrule
    \diagbox[width=12em]{Architecture}{Metrics}
              &    \MSE    &   \SSIM    &    \FID    &   \SWD
              &    \MSE    &   \SSIM    &    \FID    &   \SWD     & \tb{Speed}   \\
    \midrule
     Base
              &    0.071   &   0.522    &   16.43    &   13.02       
              &    0.119   &   0.370    &   12.06    &   25.15    & 0.0042       \\
    \hline
     ResNet-18  
              &    0.069   &   0.526    &    16.11   &   9.50      
              &    0.116   &   0.371    &    12.18   &   24.09    & 0.0045       \\
    \hline
     ResNet-34  
              &    0.066   &    0.531   & \tb{15.76} &   7.96 
              &    0.113   &    0.373   &  11.18     & \tb{22.89} & 0.0068    \\
    \hline
     ResNet-50
              & \tb{0.063} & \tb{0.539} &   15.81    &   \tb{7.31}
              & \tb{0.108} & \tb{0.380} & \tb{10.97} &   23.21    & 0.0081            \\
    \hline
    MobileNet  
              &   0.077    &   0.502    &   17.11    &   13.43
              &   0.121    &   0.362    &   13.78    &   26.13    & \tb{0.0041}   \\
    \bottomrule
  \end{tabular}
  \vspace{-5pt}
\end{table*}

\section{Encoder Analysis}\label{sec:archtecture-exploring}
In our methodology, as depicted in \cref{fig:framework}(a), we update both the encoder and discriminator while keeping the generator frozen during the training process, where the discriminator directly comes from StyleGAN \cite{stylegan}.
Hence, in this section, we analyze the impact of different neural architectures of the encoder on the quality and speed of GAN inversion.
The outcome of this analysis led us to propose a straightforward solution that enables our method to perform effectively on StyleGAN2 \cite{stylegan2}

\subsection{Encoder Architectures}\label{subsec:encoder-architecture-explore}
The encoder architecture utilized in \cref{sec:evaluation-metrics-and-applications} is a 14-layer CNN that comprises convolutional and pooling layers, culminating in a fully connected layer that produces the final latent code.
Therefore, we are interested in exploring whether modifying the encoder architecture can enhance the results of GAN inversion.
A deeper network architecture offers a higher network capacity, which can potentially improve the accuracy of approximating the inversion of the generator. 
In our analysis of different encoder architectures, we select ResNet \cite{resnet} variants (ResNet-18, ResNet-34, and ResNet-50) and MobileNet \cite{howard2017mobilenets} as backbones. 
It is important to note that the shallowest ResNet architecture has 18 layers, which is deeper than our simple encoder architecture. 
Additionally, we keep the last fully connected layer and training loss functions constantly across all the different encoder architectures.

As shown in \cref{tab:encoder-arch}, for the face, the MSE, SSIM, and SWD improve when the depth of the network increases while the FID fluctuates.
For the church, the MSE, SSIM, and FID slightly improve when the depth of the network increases while the SWD fluctuates.
However, note that the inversion time grows rapidly when the depth increases.
Regarding the inversion speed, we also experimented MobileNet \cite{howard2017mobilenets}, and the results are also reported in \cref{tab:encoder-arch}.
From the results, we could observe that the inversion speed is not accelerated obviously, but the inversion quality degenerates on some level compared to our simple encoder architecture.
This might be because MobileNet is designed for the CPU while our testing is conducted on the GPU.

\begin{figure}[!ht]
  \centering
  \includegraphics[width=0.95\linewidth]{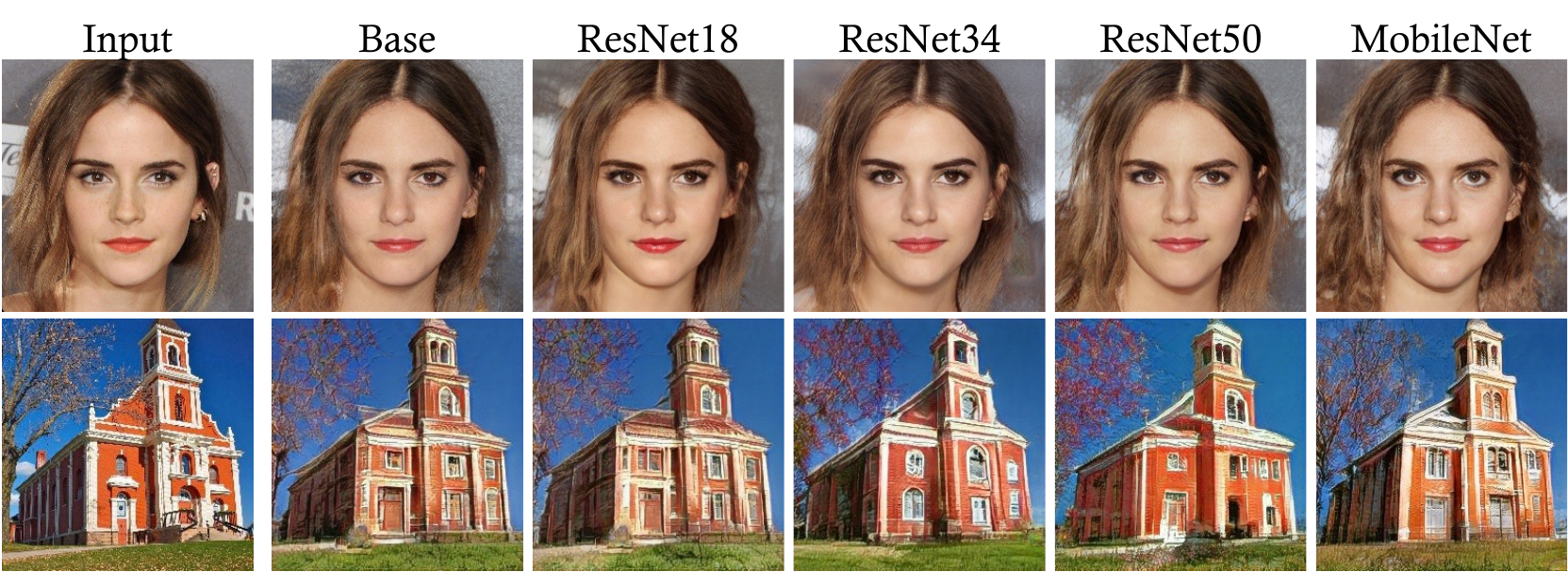}
  \vspace{-5pt}
  \caption{The qualitative reconstruction results using different network architectures on FFHQ and Church datasets.}
  \label{fig:rec-diff-arch}
  \vspace{-10pt}
\end{figure}

Based on the experiments conducted, we can conclude that increasing the depth of the encoder network generally results in improved inversion quality, as indicated by most of the performance metrics.
However, this improvement comes at the expense of a slower inversion speed, highlighting the trade-off between effectiveness and efficiency.
Nonetheless, our findings suggest that our simple encoder architecture strikes a reasonable balance between effectiveness and efficiency for StyleGAN \cite{stylegan} inversion tasks.
The visual results of the reconstruction on various architectures are included in \cref{fig:rec-diff-arch}, which also indicate that there is a minimal discernible difference between the reconstruction outputs using different architectures.
For other solutions to improve the inversion quality, a detailed discussion can be seen in \cref{sec:trade-off}.

\begin{figure}[t]
  \centering
  \includegraphics[width=0.75\linewidth]{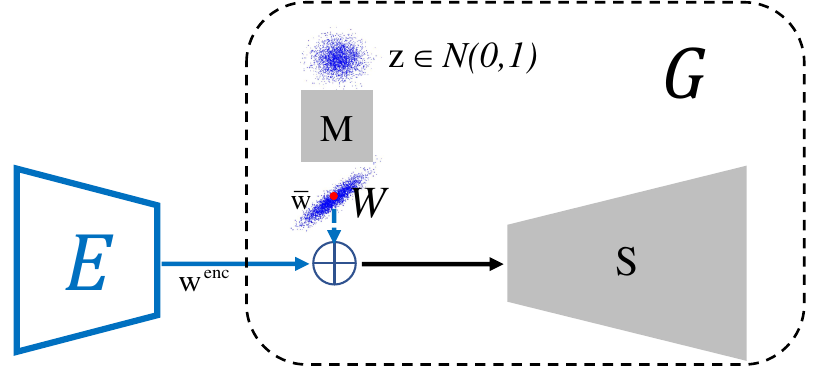}
  \vspace{-7pt}
  \caption{
    Illustration of adding average when training the encoder. The output of the encoder $\mathbf{w^{enc}}$ is firstly added to the mean value $ \bar{\tb{w}} $ from $ \mathcal{W} $, and then fed to the synthesis network. M and S are the mapping and synthesis networks in StyleGAN, respectively.
  }
  \label{fig:Framework-add-avg}
  \vspace{-10pt}
\end{figure}

\subsection{Initial Inversion Space}\label{subsec:add-average}
During the implementation of our method on StyleGAN2 \cite{stylegan2}, we encountered instability issues when training the encoder.
For instance, we observed instances where the training did not converge, indicated by the lack of loss decrease, or where the final results showed poor quality despite a decline in training loss. 
Initially, it was suspected that the use of a simple encoder architecture may have been the cause of these issues.
However, further experimentation with different encoder architectures, as discussed above, did not yield better results.
Therefore, it is possible that other factors may be contributing to the instability during training.

Recall our pipeline to train the encoder in \cref{fig:framework}(a), and it can be further illustrated with more details, as shown in \cref{fig:Framework-add-avg}, from which we can divide this pipeline into two parts.
The first one is the pre-trained generator, in which the mapping network $M$ maps a normal distribution $ \mathcal{N} $ to an unknown distribution $ \mathcal{W}_{f}$
%
Then the synthesis network $ S $ maps $ \mathcal{W}_{f} $ to the image distribution $ \mathcal{X}_f $.
The second one is the encoder, which maps the real image distribution $ \mathcal{X}_r $ to a latent distribution $ \mathcal{W}_{r} $, which is then reshaped by the synthesis network to get a reconstruction: $ x^{rec} = S(E(x^{real})) $.
When the encoder training is finished, we hope the distribution produced by the encoder $ \mathcal{W}_{r} $ could align well with the distribution generated by the mapping network $ \mathcal{W}_{f} $ no matter how far away those two distributions are in the initial state.
Also, remember that the generator is fixed when the encoder is training, which means that $ \mathcal{W}_{f} $ is still and stationary.
As a result, the process of training the encoder aims to eventually align its output $ \mathcal{W}_{r} $ to $ \mathcal{W}_{f} $.
Intuitively, when these two distributions are close at the beginning of training, it is stable to train the encoder; otherwise, when they are far from each other, the training can become unstable, especially with StyleGAN2.
To address this issue, we can improve the initial alignment of these two distributions by adding the mean value of $ \mathcal{W}_{f} $ to the initial distribution of $ \mathcal{W}_{r} $. 
%
%
To be specific, the output of the encoder is firstly added with the average value of $ \mathcal{W}_{f} $ and then fed to the synthesis network:
\begin{align}
  x^{rec} = S(E(x^{real}) + \bar{\bf{w}}), \label{eq:add_w_avg}
\end{align}
where $ \bar{\tb{w}} = {\E}_{\z \sim P(\z)}[M(\z)] $ is the mean value of $ \mathcal{W}_{f} $.
There are three advantages to adding such an average, as listed below.
Firstly, from an optimization perspective, adding $ \bar{\tb{w}} $ to $ \mathcal{W}_{r} $ ensures that these two distributions are initially close to each other, easing the optimization process and accelerating the convergence speed.
Secondly, when $ \bar{\tb{w}} $ is added, the encoder only needs to learn the residual information for each input image, reducing the learning difficulty.
Third, $ \bar{\tb{w}} $ can be viewed as a regularizer that forces the output values from the encoder to be closer to the average $ \bar{\tb{w}} $, promoting the generation of more in-domain latent codes.
It is noted that when synthesizing an image in StyleGAN \cite{stylegan}, a given $ \tb{w} $ is truncated by the $ \bar{\tb{w}} $ using the linear interpolation $\tb{w}' = \bar{\tb{w}} + \psi (\tb{w} - \bar{\tb{w}})$, where $\psi < 1$.
This truncation trick aims at improving the average image quality with some degree of degradation in the variation of the generated images.
Instead, the purpose of adding the $ \bar{\tb{w}} $ in our pipeline is to ease the encoder learning.

\setlength{\tabcolsep}{4.0pt}
\begin{table}[t]
  \setlength{\tabcolsep}{3.0pt}
  \caption{ 
  The inversion results whether or not adding w average when training the encoder on StyleGAN2 on the FFHQ dataset, in which we could see that the inversion results significantly improved from all metrics when adding the average to the encoder. Besides, we also report the results from e4e \cite{tov2021designing}.
  }
  \label{tab:compare-avg}
  \vspace{-8pt}
  \scriptsize\centering
  \begin{tabular}{l|cccc|cc}
                    \toprule
                    & \multicolumn{4}{c|}{Inversion}
                    & \multicolumn{2}{c}{Interpolation}                                           \\
                    \midrule
       Metrics      &    \FID    &    \SWD    &    \MSE    &  \SSIM     &    \FID    &  \SWD      \\
                    \midrule
       e4e \cite{tov2021designing}
                    &    39.48   &    12.97   &    0.054   &  0.562     &    83.59   & \tb{34.48} \\
                    \midrule

       \emph{w/o} Average 
                    &   27.95    &    22.46   &    0.074   &   0.529    &    70.69   &  38.15     \\
                    \bottomrule
       \emph{w/} Average  
                    & \bf{14.93} &  \tb{8.37} & \bf{0.036} & \bf{0.585} & \tb{52.52} &  36.71     \\
                    \midrule
 \end{tabular}
  \vspace{-10pt}
\end{table}

\begin{figure}[t]
  \centering
  \includegraphics[width=0.96\linewidth]{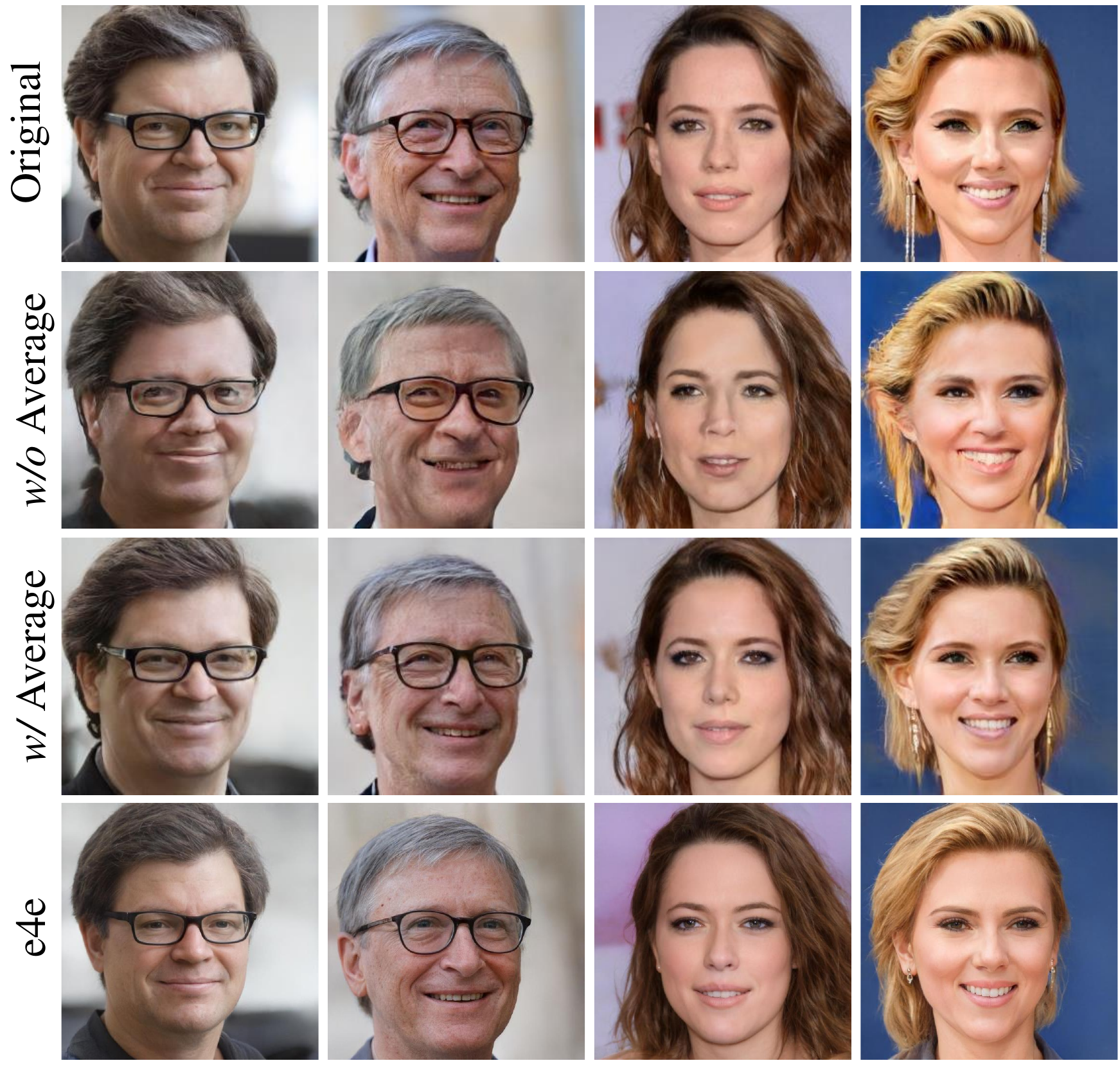}
  \vspace{-15pt}
  \caption{
    Comparison results on encoder reconstruction whether the average is added or not when training encoders. We also add the comparison results with e4e \cite{tov2021designing} in the last row. Zoom in for the details.
  }
  \label{fig:w/o-avg}
  \vspace{-10pt}
\end{figure}

We conducted a comparison study to evaluate the impact of adding $ \bar{\tb{w}} $ on StyleGAN2~\cite{stylegan2}.
As previously mentioned, the encoder training process may fail (\textit{e.g.}, by diverging loss) when $ \bar{\tb{w}} $ is not included.
Therefore, we selected a case where the loss was observed to decrease normally, as demonstrated by the training loss curves in \cref{fig:noise-level-average}(b). 
Note that both loss curves are plotted on a logarithmic scale, as the loss magnitude without the average is significantly larger than the loss with the average, making it difficult to compare.
The training loss without $\bar{\tb{w}}$ is consistently much larger than the loss with $\bar{\tb{w}}$ throughout the training process, as shown in \cref{fig:noise-level-average}(b).
Moreover, the encoder reconstruction results for both cases, with and without $\bar{\tb{w}}$, are presented in \cref{fig:w/o-avg}, and it is clear that the encoder reconstruction quality significantly improves when adding the average.
Quantitative results, presented in \cref{tab:compare-avg}, further support the efficacy of adding the average in improving the performance of the encoder.
In \cref{sec:vis-initial-inversion-space}, we present visualized results of the initial reconstruction with and without the inclusion of the average latent code during encoder training.
Moreover, we compared our method with e4e \cite{tov2021designing} since it also trained on StyleGAN2 \cite{stylegan2}.
As shown in \cref{fig:w/o-avg}, the results from e4e show less fidelity compared to our method with $\bar{\tb{w}}$ added.
\cref{tab:compare-avg} also demonstrated that our encoder with  $\bar{\tb{w}}$ added could beat e4e in most metrics except SWD in the interpolation task.

\section{Trade-off between Reconstruction Quality and Editability} \label{sec:trade-off}
This section first gives an ablation study on the hyper-parameter $ \lambda_{dom} $ in \cref{eq:optimization}, from which we observe a trade-off on the inverted code between its ability to reconstruct an input image and its ability to be well-edited.
To further verify this observation, we conducted two additional experiments. 
Firstly, we retrained the generator using an enlarged $W$ space. 
Secondly, we extended the inversion space by optimizing both the latent codes and noise maps.
Noticeably, some work also finds this trade-off.
For instance, e4e \cite{tov2021designing} analyzes the structure of StyleGAN's latent space and tries to land the inverted latent codes in the original domain of the StyleGAN as \cite{zhu2020indomain} dose.
However, here we demonstrate this trade-off from another perspective, i.e., dimension.

\subsection{Ablation Study}\label{subsec:ablation-study}
\begin{figure}[t]
  \centering
  \includegraphics[width=0.70\linewidth]{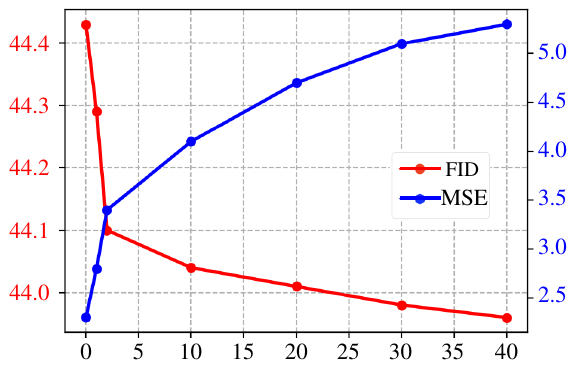}
  \vspace{-10pt}
  \caption{The quantitative results on $ \lambda_{dom} $ when doing optimization using \cref{eq:optimization}. The $ x $-axis indicates the value of $ \lambda_{dom} $. For the MSE (blue line), the value is re-scaled by 100.
    }
  \label{fig:lambda}
\end{figure}

\begin{figure}[t]
  \centering
  \includegraphics[width=1.0\linewidth]{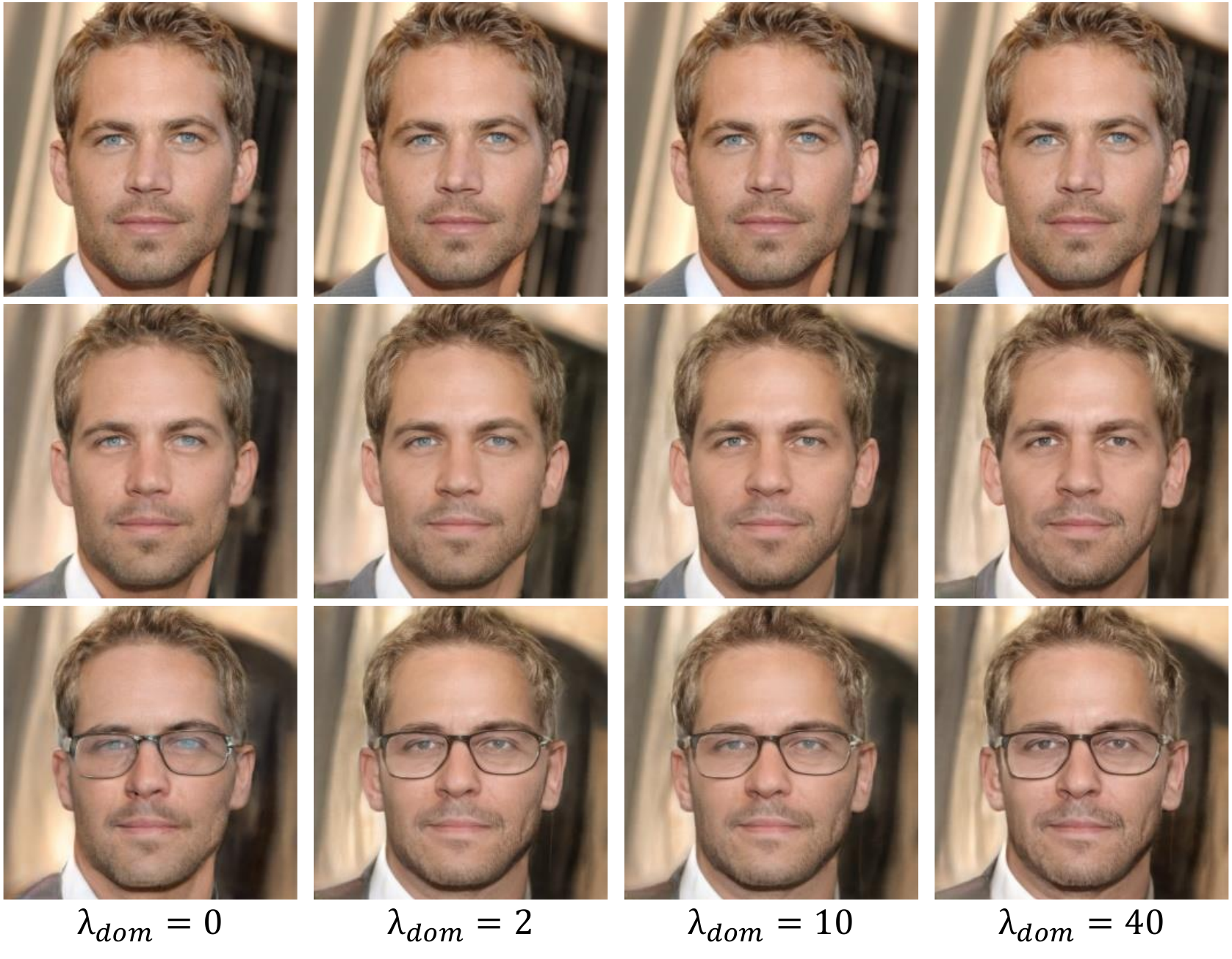}
  \vspace{-22pt}
  \caption{
    Ablation study on the loss weight in \cref{eq:optimization} for the \emph{domain-regularized} optimization.
    From top to bottom: original images, reconstructed images, and manipulation results (wearing eyeglasses).
    For each group of images, the weight $\lambda_{dom}$ is set to be 0, 2, 40.
    When $\lambda_{dom}$ equals to 0, it produces the best reconstructed results but relatively poor manipulation results.
    When $\lambda_{dom}$ equals 40, we get worse reconstruction but more satisfying manipulation.
  }
  \label{fig:encoder-reg-weight}
  \vspace{-15pt}
\end{figure}

As described in \cref{sec:method}, after the initial training of the encoder, we perform the \emph{domain-regularized} optimization on each image to further improve the reconstruction quality.
Different from the previous MSE-based optimization, we involve the learned \emph{domain-guided} encoder as a regularizer to land the inverted code inside the semantic domain, as described in \cref{eq:optimization}.
Here, we study the role of the encoder in the optimization process by varying the weight $\lambda_{dom}$ in \cref{eq:optimization}.
\cref{fig:lambda} reports the quantitative results on $ \lambda_{dom} $ when inverting (MSE) and manipulating (FID) the corresponding images, in which we could observe that with the increasing of $ \lambda_{dom} $, the reconstruction MSE is growing while the FID of the manipulated images is decreasing.
This indicates a trade-off between the image reconstruction precision and the editing property.
\cref{fig:encoder-reg-weight} gives qualitative comparison between $\lambda_{dom} = 0, 2, 10, 40$, in which we also observe this trade-off.
a larger $\lambda_{dom}$ will bias the optimization towards the domain constraint such that the inverted codes are more semantically meaningful (e.g., semantics will have a better alignment with the encoder).
Instead, the cost is that the target image cannot be ideally recovered for per-pixel values.
In practice, we set $\lambda_{dom} = 2$.

\subsection{Retraining Generators}\label{subsec:retraining}
In this part, we investigate whether retraining the generator, as done in \cite{zhu2020indomain}, is necessary.
Recall that the generator of StyleGAN has two parts.
The first part is the mapping network, which consists of eight fully connected layers followed by Leaky ReLU, that maps a latent code $ \z $ to an intermediate latent code $ \w $, and the space formed by those intermediate latent codes is called $ \W $ space.
Both the dimension of $ \z $ and $ \w $ is 512 in StyleGAN.
The other part of the generator is called the synthesis network, which takes the $ \w $ as input in a layer-wise manner via the Adaptive Instance Normalization (AdaIN) \cite{adain} operator and generates synthesized images. 
For a generator that could synthesize $ 256 \times 256 $ images, the synthesis network contains 14 convolution layers.
That means the single latent code $ \w $ produced by the mapping network will be repeated 14 times and then fed to each convolution layer.
For the GAN inversion task,  some work \cite{image2stylegan, zhu2020indomain} introduced the $ \WP $ space.
Namely, they use 14 different $ \w $s to reconstruct an image instead of using a single $ \w $ but repeat 14 times.
That means the dimension of the inverted codes is enlarged by 14 times (i.e., the dimension of the $ \W $ space is 512, and the dimension of $ \WP $ is $ 14 \times 512 = 7168 $) than the original one, and the inversion results have improved substantially.
To be consistent with the $ \WP $ space, Zhu \emph{et al.} \cite{zhu2020indomain} changed the output layer of the mapping network, i.e., the dimension of the output layer is changed to $ 14 \times 512 $.
One disadvantage of this is the generator needs to be retrained, which is time-consuming.
Hence, in this section, we explore whether we do not need to retrain the generator to use  $ \WP $ to invert an image.
%

\begin{figure}[t]
  \centering
  \includegraphics[width=0.95\linewidth]{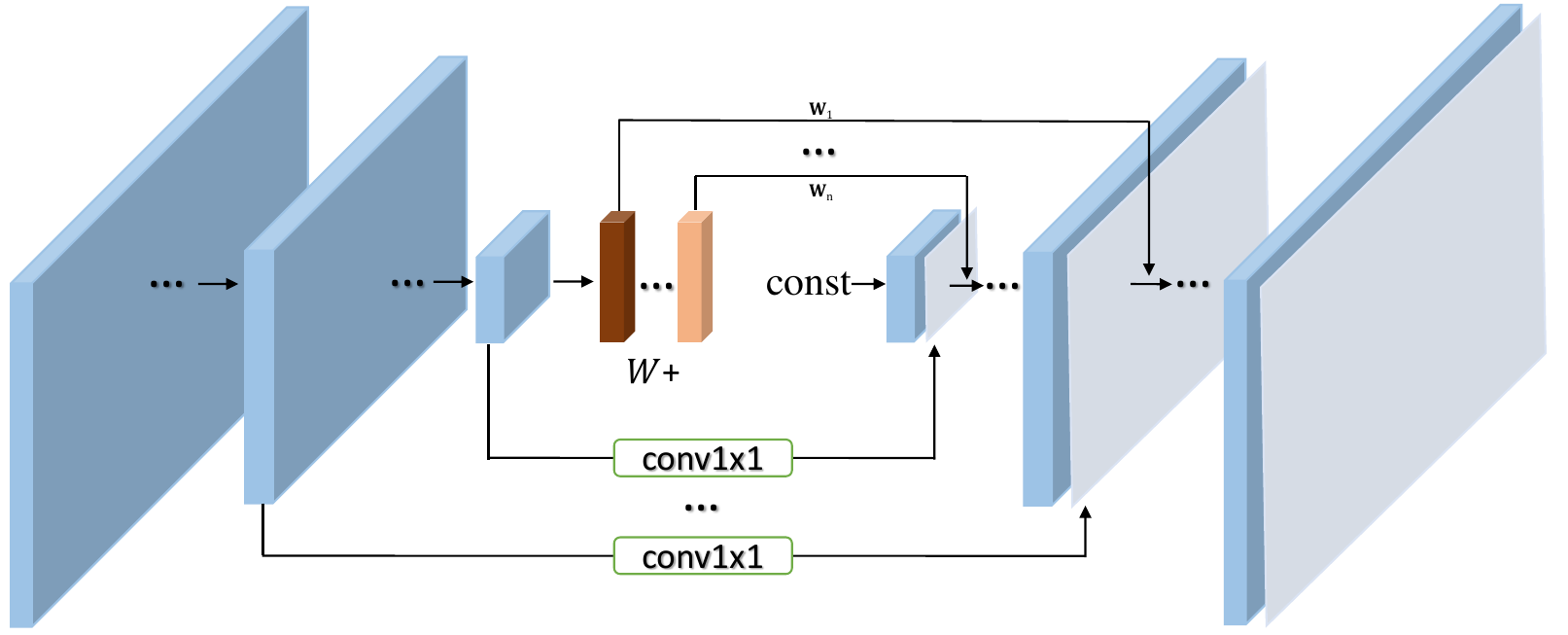}
  \vspace{-5pt}
  \caption{
    The framework of injecting noises is able to extract both the latent representation and the prior of the input image. The encoder is designed symmetrically with the generator to align with its noises. And for the feature maps of the encoder having the same resolution as feature maps of the generator, we use a $1 \times 1$ convolution to compress the channel dimension to 1 and then feed to the generator.
  }
  \label{fig:Framework-noise}
  \vspace{-10pt}
\end{figure}

\cref{tab:compare-retraining} presents the quantitative results for reconstruction and interpolation on FFHQ using StyleGAN and StyleGAN2 with or without retraining the generator.
As shown, we observed that the reconstruction improves when the generator is retrained, while the interpolation quality deteriorates.
Conversely, when the generator is not retrained, the interpolation quality improves, while the reconstruction quality deteriorates.
This is reasonable because when retraining the generator, the $ \W $ space of the generator is 14 times larger than that of the generator without retraining, providing a larger space for the encoder to explore and naturally benefiting the reconstruction quality. 
However, the larger space also means that the same semantics in one image are spread into a larger space, leading to a dilution of those semantics. 
Consequently, those scattered semantics are challenging to align well on some level and result in relatively poor interpolation results.
In summary, these results provide further evidence for the trade-off between reconstruction quality and editing properties.

We noticed from~\cref{tab:compare-retraining} that the reconstruction results for StyleGAN2 are better than those for StyleGAN, but the interpolation results are just the opposite.
We suspect that this discrepancy is due to the statistical average code $\bar{\w}$. 
All inverted codes are built on top of $\bar{\w}$, which restricts the diversity when interpolating two codes, ultimately harming the FID metric.
We conducted experiments on other datasets, such as LSUN Tower and LSUN Bedroom, and the results, shown in \cref{tab:val-on-bedroom-tower}, confirm the pattern observed in \cref{tab:compare-retraining}.

\setlength{\tabcolsep}{5.0pt}
\begin{table*}[t]
  \setlength{\tabcolsep}{5.5pt}
  \caption{
          The quantitative results on whether or not to retrain the generator when training the encoder on StyleGAN and StyleGAN2 on FFHQ dataset.
  }
  \label{tab:compare-retraining}
  \vspace{-8pt}
  \scriptsize\centering
  \begin{tabular}{l|cccc|cc|cccc|cc}
    \toprule
      &  \multicolumn{6}{c|}{StyleGAN \cite{stylegan}}
      &  \multicolumn{6}{c}{StyleGAN2 \cite{stylegan2}} \\
    \midrule
      &  \multicolumn{4}{c|}{Reconstruction}
      &  \multicolumn{2}{c|}{Interpolation}
      &  \multicolumn{4}{c|}{Reconstruction}
      &  \multicolumn{2}{c}{Interpolation}    \\
    \midrule
 Metrics
        &    \FID    &    \SWD    &     \MSE    &    \SSIM     &    \FID     &   \SWD
        &    \FID    &    \SWD    &     \MSE    &    \SSIM     &    \FID     &   \SWD       \\
      \midrule

 \emph{w/o} Retraining
        &  16.08     &    17.34   &    0.061    &    0.551     & \tb{44.97}  & \tb{25.38}
        &  15.30     &    9.28    &    0.049    &    0.575     & \tb{50.46}  & \tb{29.60}   \\
 \emph{w/} Retraining
        & \tb{15.07} & \tb{14.95} & \tb{0.053}  & \tb{0.568}   &    46.56    &  25.66
        & \tb{14.93} & \tb{8.37}  & \tb{0.036}  & \tb{0.585}   &    52.52    &  36.71       \\
      \bottomrule
  \end{tabular}
  \vspace{-5pt}
\end{table*}

\subsection{Extending the Inversion Space}\label{subsec:extending-space}
As seen in \cref{subsec:retraining}, the dimension of the latent space can impact the reconstruction results. 
So any other ways to involve more dimensions to improve the reconstruction precision further? 
The answer is yes!
Previous work \cite{bartz2020one,stylegan2,image2stylegan++} has explored the noises in the feature maps of the generator when inverting an image. 
Therefore, in this part, we attempt to incorporate these noises into our proposed encoder.
%
Specifically, the StyleGAN-based generators inject random noise into the feature maps to improve both the quality and stochastic variations of synthesized images. 
Those stochastic variations, for example, include the change in the hair, silhouettes, and background when using the same latent code but resampling the noise in the feature maps.
When treating these generators as fixed decoders during the encoder training process, the random noise must be fixed since each input corresponds to one output. 
Hence, we hope those noises can be utilized when reconstructing the input image instead of fixing them.
Namely, the random noises in the generator will be replaced by the features extracted from the encoder, as shown in \cref{fig:Framework-noise}.

In this scenario, the encoder is symmetrically designed with the generator to match the number and the resolution of the noise maps.
Take the $ 256 \times 256 $ images as an example. 
There are fourteen layers in the generator, which include seven blocks (i.e., from resolution $ 4 \times 4 $ to $256 \times 256 $, and for each block, there are two convolutional layers).
To simplify notation, we use B-1 to B-7 to represent blocks 1 to 7.
%
We gradually replaced the noises in the generator to observe their influence on the reconstruction results.
%
Concretely, we first replace the noises in B-1 using the features extracted from the encoder at resolution $ 4 \times 4 $, while the remaining blocks still use fixed noises. 
Then, we add B-2 by replacing the noises in the generator with the features extracted from the encoder at resolution $ 8 \times 8 $. 
Finally, B-7 means replacing all noises in the generator with the features extracted from the encoder.

\begin{figure}[t]
  \centering
  \includegraphics[width=0.96\linewidth]{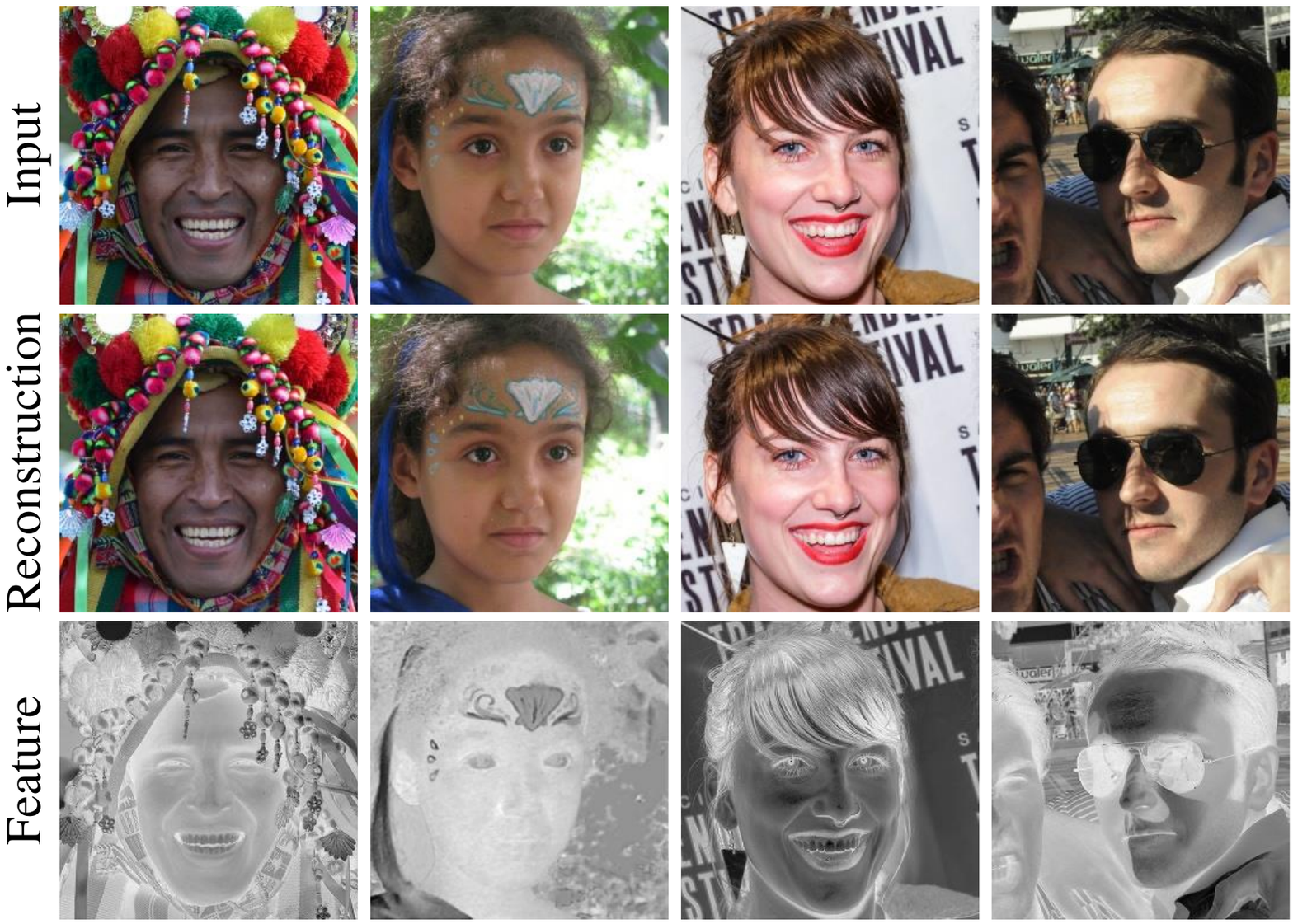}
  \vspace{-10pt}
  \caption{
    The encoder reconstruction when using all noises in the generator. The last row shows the features extracted from the encoder at the resolution $ 256 \times 256$.
  }
  \label{fig:noise-feats}
  \vspace{-10pt}
\end{figure}

\begin{figure}[t]
  \centering
  \includegraphics[width=0.96\linewidth]{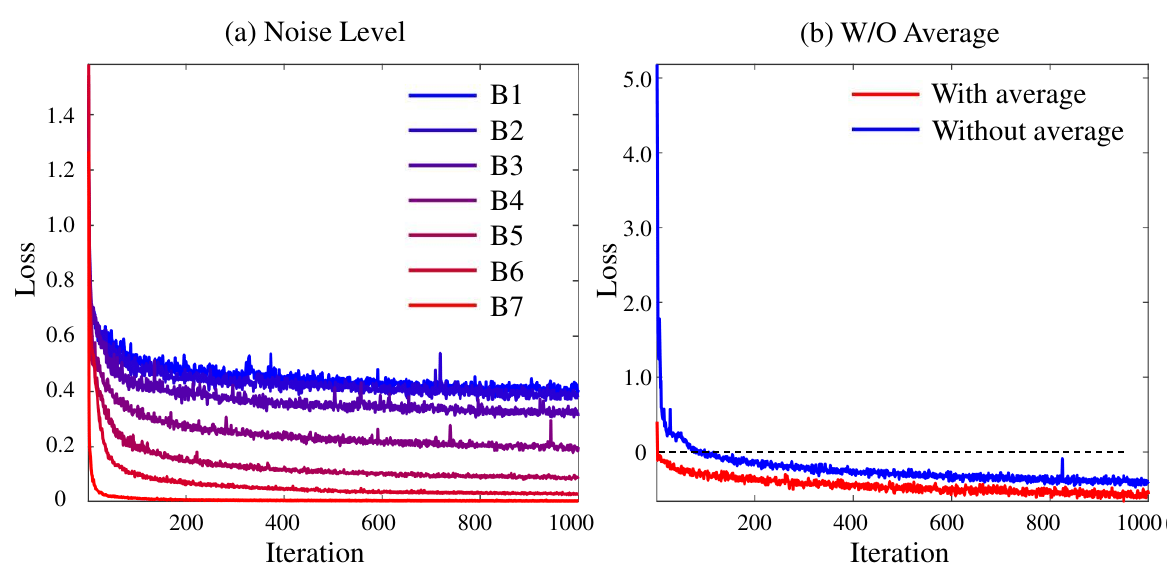}
  \vspace{-10pt}
  \caption{
    Loss curve during training. (a) is the loss curve when involving noises at different blocks. (b) shows the loss curve whether or not adding the average when training the encoder on StyleGAN2.
    }
  \label{fig:noise-level-average}
  \vspace{-15pt}
\end{figure}

In Fig.~\ref{fig:noise-level-average}(a), we present the training loss on the FFHQ dataset when replacing the corresponding noise at each block, in which we see that with more and more noises being replaced, the training loss becomes lower and lower.
Notably, when all noises are replaced, the training loss becomes minimal at the very beginning, taking only half an hour on 8 RTX 2080 Ti GPUs.
\cref{tab:encoder-noise} shows the quantitative results, in which we report the MSE and SSIM on the reconstruction and FID and SWD on the interpolation when involving different blocks.
We observe from the \cref{tab:encoder-noise} that the reconstruction MSE decreases more than one hundred times when replacing all noise in the generator, and the SSIM also improves nearly one time.
However, the performance on the editing task, such as interpolation, worsens as more noises are involved, providing further evidence of the existence of the trade-off.
\cref{fig:noise-feats} shows some reconstruction results as well as the features extracted by the encoder at the first convolution layer of B-7, which shows that the very complicated images could be recovered, including their backgrounds and many other details.
This is sensible since when replacing all noises in the generator, the encoder will use 181,920 parameters to reconstruct an image of size $ 256 \times 256$, which has 196,608 pixels in total.
One of the possible directions could be image processing since its U-Net-like architecture could precisely reconstruct the input but with lower network capacity compared to the original U-Net \cite{unet}.
However, we do not discuss this further since our paper focuses on image editing.

In conclusion, with the improvement of inversion precision, the editing quality tends to decrease. 
This is normal because the generator cannot synthesize an image that perfectly matches the real images from the latent space. 
Therefore, with the improvement of the reconstruction precision on real images, some extra information needs to be borrowed or activated, such as noise in our case, which plays an assistant role when synthesizing an image but plays an essential role in reconstructing a real image perfectly.
Thus, when reusing the knowledge learned by GANs on real images, an apparent deterioration emerges.
Hence, a trade-off must exist between reconstruction precision and editing quality.

\setlength{\tabcolsep}{6.5pt}
\begin{table}[t]
  \setlength{\tabcolsep}{4.5pt}
  \caption{
     Comparison results on replacing the noise in the generator at different blocks on the FFHQ dataset.
     }
  \label{tab:encoder-noise}
  \vspace{-8pt}
  \scriptsize\centering
  \begin{tabular}{l|ccccccc}
    \toprule
            &   B-1   &  B-2   &   B-3   &   B-4   &   B-5   &   B-6   &  B-7      \\
    \hline
    \MSE    &  0.068  & 0.065  &  0.047  &  0.025  &  0.010  &  0.004  &  0.00038  \\
    \SSIM   &  0.495  & 0.498  &  0.531  &  0.634  &  0.765  & 0.886   &  0.989    \\
    \FID    &  41.89  & 43.39  &  44.29  &  48.74  &  49.58  & 52.89   &  58.26    \\
    \SWD    &  59.91  & 62.67  &  65.44  &  65.48  & 65.08   & 65.56  &  66.49    \\
    \bottomrule
  \end{tabular}
  \vspace{-5pt}
\end{table}

\section{Discussion and Conclusion}\label{sec:conclusion}
\begin{figure}[t]
  \centering
  \includegraphics[width=1\linewidth]{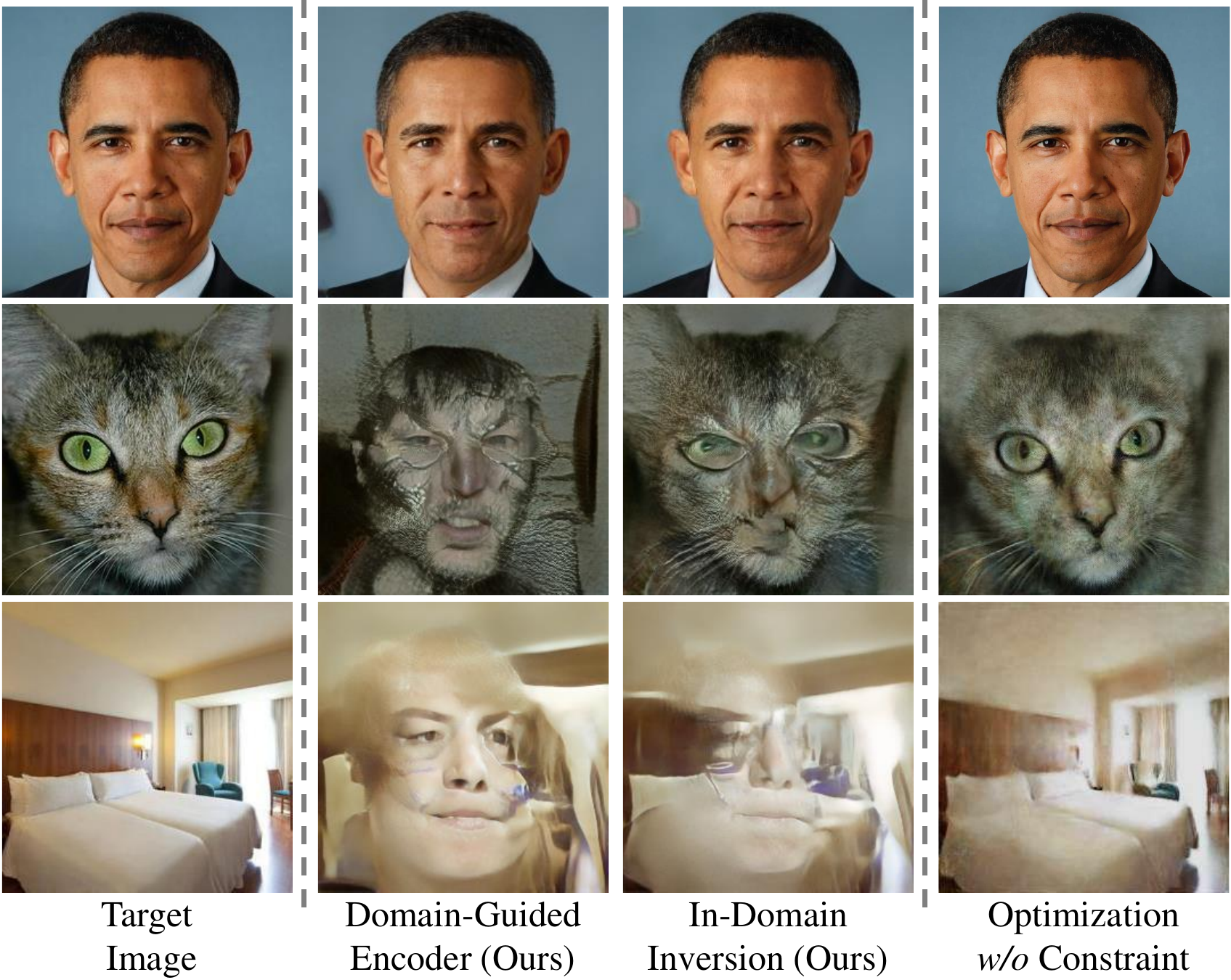}
  \vspace{-15pt}
  \caption{
    Results on the inverting face, cat face, and bedroom using the same face synthesis model.
    From left to right: target images, reconstruction results with the outputs from the \emph{domain-guided} encoder, reconstruction results with the proposed \emph{in-domain} inversion, reconstruction results by directly optimizing the latent code \emph{w/o} considering domain alignment \cite{image2stylegan}.
  }
  \label{fig:cross-domain}
  \vspace{-10pt}
\end{figure}

In this work, we explore the \emph{semantic} property of the inverted codes in the GAN inversion task and propose a novel \emph{in-domain} inversion method.
To the best of our knowledge, this is the first attempt to invert a pre-trained GAN model \emph{explicitly} considering the semantic knowledge encoded in the latent space.
We show that the code simply recovering the pixel value of the target image is not sufficient to represent the image at the semantic level.
For example, in \cref{fig:cross-domain}, we invert different types of image instances (\emph{i.e.}, face, cat face, and bedroom) with the face synthesis model.
The last column shows the results from Image2StyleGAN \cite{image2stylegan}, which recovers a cat or a bedroom with the domain knowledge learned to synthesize human faces.
By contrast, the face outline can still be observed in the reconstructions using our \emph{in-domain} inversion (third column).
This demonstrates, from a different angle, the superiority of our approach in producing semantically meaningful codes.
Take inverting bedroom (third row) as an example. 
The bedroom image is outside the domain of the training data, and the GAN model should not be able to learn the bedroom-related semantics.
Accordingly, reusing the face knowledge to represent a bedroom is ill-defined.
Even though we can always use more parameters to over-fit the pixel values of the bedroom (\emph{e.g.}, the last column), such over-fitting would fail to support semantic image manipulation.
From this viewpoint, our \emph{in-domain} inversion lands the inverted code inside the original domain to make it semantically meaningful.
In other words, we aim at finding an adequate code to recover the target image from \emph{both the pixel level and the semantic level}.
Such \emph{in-domain} inversion significantly facilitates real image editing.

\noindent\textbf{Acknowledgement}: The project is partially supported by the Amazon Research Award.

\ifCLASSOPTIONcaptionsoff
  \newpage
\fi
\bibliographystyle{IEEEtran}
\bibliography{ref}

\begin{thebibliography}{10}
\providecommand{\url}[1]{#1}
\csname url@samestyle\endcsname
\providecommand{\newblock}{\relax}
\providecommand{\bibinfo}[2]{#2}
\providecommand{\BIBentrySTDinterwordspacing}{\spaceskip=0pt\relax}
\providecommand{\BIBentryALTinterwordstretchfactor}{4}
\providecommand{\BIBentryALTinterwordspacing}{\spaceskip=\fontdimen2\font plus
\BIBentryALTinterwordstretchfactor\fontdimen3\font minus \fontdimen4\font\relax}
\providecommand{\BIBforeignlanguage}[2]{{%
\expandafter\ifx\csname l@#1\endcsname\relax
\typeout{** WARNING: IEEEtran.bst: No hyphenation pattern has been}%
\typeout{** loaded for the language `#1'. Using the pattern for}%
\typeout{** the default language instead.}%
\else
\language=\csname l@#1\endcsname
\fi
#2}}
\providecommand{\BIBdecl}{\relax}
\BIBdecl

\bibitem{pggan}
T.~Karras, T.~Aila, S.~Laine, and J.~Lehtinen, ``Progressive growing of {GAN}s for improved quality, stability, and variation,'' in \emph{ICLR}, 2018.

\bibitem{stylegan}
T.~Karras, S.~Laine, and T.~Aila, ``A style-based generator architecture for generative adversarial networks,'' in \emph{CVPR}, 2019.

\bibitem{stylegan2}
T.~Karras, S.~Laine, M.~Aittala, J.~Hellsten, J.~Lehtinen, and T.~Aila, ``Analyzing and improving the image quality of {StyleGAN},'' in \emph{CVPR}, 2020.

\bibitem{gan}
I.~Goodfellow, J.~Pouget-Abadie, M.~Mirza, B.~Xu, D.~Warde-Farley, S.~Ozair, A.~Courville, and Y.~Bengio, ``Generative adversarial nets,'' in \emph{NeurIPS}, 2014.

\bibitem{goetschalckx2019ganalyze}
L.~Goetschalckx, A.~Andonian, A.~Oliva, and P.~Isola, ``{GANalyze}: Toward visual definitions of cognitive image properties,'' in \emph{ICCV}, 2019.

\bibitem{gansteerability}
A.~Jahanian, L.~Chai, and P.~Isola, ``On the "steerability" of generative adversarial networks,'' \emph{arXiv preprint arXiv:1907.07171}, 2019.

\bibitem{shen2020interfacegan}
Y.~Shen, C.~Yang, X.~Tang, and B.~Zhou, ``{InterFaceGAN}: Interpreting the disentangled face representation learned by {GAN}s,'' \emph{TPAMI}, 2020.

\bibitem{shen2021closed}
Y.~Shen and B.~Zhou, ``Closed-form factorization of latent semantics in {GANs},'' in \emph{CVPR}, 2021.

\bibitem{zhu2016generative}
J.-Y. Zhu, P.~Kr{\"a}henb{\"u}hl, E.~Shechtman, and A.~A. Efros, ``Generative visual manipulation on the natural image manifold,'' in \emph{ECCV}, 2016.

\bibitem{luo2017learning}
J.~Luo, Y.~Xu, C.~Tang, and J.~Lv, ``Learning inverse mapping by autoencoder based generative adversarial nets,'' in \emph{ICNIP}, 2017.

\bibitem{lipton2017precise}
Z.~C. Lipton and S.~Tripathi, ``Precise recovery of latent vectors from generative adversarial networks,'' in \emph{ICLR Workshop}, 2017.

\bibitem{creswell2018inverting}
A.~Creswell and A.~A. Bharath, ``Inverting the generator of a generative adversarial network,'' \emph{TNNLS}, 2018.

\bibitem{bau2019inverting}
D.~Bau, J.-Y. Zhu, J.~Wulff, W.~Peebles, H.~Strobelt, B.~Zhou, and A.~Torralba, ``Inverting layers of a large generator,'' in \emph{ICLR Workshop}, 2019.

\bibitem{lia}
J.~Zhu, D.~Zhao, B.~Zhang, and B.~Zhou, ``Disentangled inference for {GAN}s with latently invertible autoencoder,'' \emph{arXiv preprint arXiv:1906.08090}, 2019.

\bibitem{invertibility}
F.~Ma, U.~Ayaz, and S.~Karaman, ``Invertibility of convolutional generative networks from partial measurements,'' in \emph{NeurIPS}, 2018.

\bibitem{image2stylegan}
A.~Rameen, Q.~Yipeng, and W.~Peter, ``{Image2StyleGAN}: How to embed images into the stylegan latent space?'' in \emph{ICCV}, 2019.

\bibitem{zhu2020indomain}
J.~Zhu, Y.~Shen, D.~Zhao, and B.~Zhou, ``In-domain {GAN} inversion for real image editing,'' in \emph{ECCV}, 2020.

\bibitem{sngan}
T.~Miyato, T.~Kataoka, M.~Koyama, and Y.~Yoshida, ``Spectral normalization for generative adversarial networks,'' in \emph{ICLR}, 2018.

\bibitem{wgan}
M.~Arjovsky, S.~Chintala, and L.~Bottou, ``Wasserstein generative adversarial networks,'' in \emph{ICML}, 2017.

\bibitem{wgan_gp}
I.~Gulrajani, F.~Ahmed, M.~Arjovsky, V.~Dumoulin, and A.~C. Courville, ``Improved training of wasserstein {GAN}s,'' in \emph{NeurIPS}, 2017.

\bibitem{whichgan}
L.~Mescheder, S.~Nowozin, and A.~Geiger, ``Which training methods for {GAN}s do actually converge?'' in \emph{ICML}, 2018.

\bibitem{biggan}
A.~Brock, J.~Donahue, and K.~Simonyan, ``Large scale {GAN} training for high fidelity natural image synthesis,'' in \emph{ICLR}, 2019.

\bibitem{stylegan2ada}
T.~Karras, M.~Aittala, J.~Hellsten, S.~Laine, J.~Lehtinen, and T.~Aila, ``Training generative adversarial networks with limited data,'' in \emph{NeurIPS}, 2020.

\bibitem{stylegan3}
T.~Karras, M.~Aittala, S.~Laine, E.~H\"ark\"onen, J.~Hellsten, J.~Lehtinen, and T.~Aila, ``Alias-free generative adversarial networks,'' in \emph{NeurIPS}, 2021.

\bibitem{dcgan}
A.~Radford, L.~Metz, and S.~Chintala, ``Unsupervised representation learning with deep convolutional generative adversarial networks,'' in \emph{ICLR}, 2016.

\bibitem{ramesh2019spectral}
A.~Ramesh, Y.~Choi, and Y.~LeCun, ``A spectral regularizer for unsupervised disentanglement,'' in \emph{ICML}, 2019.

\bibitem{plumerault2020controlling}
A.~Plumerault, H.~L. Borgne, and C.~Hudelot, ``Controlling generative models with continuous factors of variations,'' in \emph{ICLR}, 2020.

\bibitem{voynov2020unsupervised}
A.~Voynov and A.~Babenko, ``Unsupervised discovery of interpretable directions in the {GAN} latent space,'' in \emph{ICML}, 2020.

\bibitem{shen2019interpreting}
Y.~Shen, J.~Gu, X.~Tang, and B.~Zhou, ``Interpreting the latent space of {GAN}s for semantic face editing,'' in \emph{CVPR}, 2020.

\bibitem{yang2019semantic}
C.~Yang, Y.~Shen, and B.~Zhou, ``Semantic hierarchy emerges in deep generative representations for scene synthesis,'' \emph{IJCV}, 2020.

\bibitem{ganspace}
E.~H{\"a}rk{\"o}nen, A.~Hertzmann, J.~Lehtinen, and S.~Paris, ``{GANSpace}: Discovering interpretable {GAN} controls,'' in \emph{NeurIPS}, 2020.

\bibitem{zhu2021lowrankgan}
J.~Zhu, R.~Feng, Y.~Shen, D.~Zhao, Z.~Zha, J.~Zhou, and Q.~Chen, ``Low-rank subspaces in {GAN}s,'' in \emph{NeurIPS}, 2021.

\bibitem{liu2015faceattributes}
Z.~Liu, P.~Luo, X.~Wang, and X.~Tang, ``Deep learning face attributes in the wild,'' in \emph{ICCV}, 2015.

\bibitem{zhou2017places}
B.~Zhou, A.~Lapedriza, A.~Khosla, A.~Oliva, and A.~Torralba, ``Places: A 10 million image database for scene recognition,'' \emph{TPAMI}, 2017.

\bibitem{bau2019semantic}
D.~Bau, H.~Strobelt, W.~Peebles, J.~Wulff, B.~Zhou, J.-Y. Zhu, and A.~Torralba, ``Semantic photo manipulation with a generative image prior,'' \emph{SIGGRAPH}, 2019.

\bibitem{ali}
V.~Dumoulin, I.~Belghazi, B.~Poole, O.~Mastropietro, A.~Lamb, M.~Arjovsky, and A.~Courville, ``Adversarially learned inference,'' in \emph{ICLR}, 2017.

\bibitem{bigan}
J.~Donahue, P.~Kr{\"a}henb{\"u}hl, and T.~Darrell, ``Adversarial feature learning,'' in \emph{ICLR}, 2017.

\bibitem{donahue2019bigbigan}
J.~Donahue and K.~Simonyan, ``Large scale adversarial representation learning,'' in \emph{NeurIPS}, 2019.

\bibitem{perarnau2016invertible}
G.~Perarnau, J.~Van De~Weijer, B.~Raducanu, and J.~M. {\'A}lvarez, ``Invertible conditional {GAN}s for image editing,'' in \emph{NeurIPS Workshop}, 2016.

\bibitem{richardson2021encoding}
E.~Richardson, Y.~Alaluf, O.~Patashnik, Y.~Nitzan, Y.~Azar, S.~Shapiro, and D.~Cohen-Or, ``Encoding in style: a stylegan encoder for image-to-image translation,'' in \emph{CVPR}, June 2021.

\bibitem{xu2021generative}
Y.~Xu, Y.~Shen, J.~Zhu, C.~Yang, and B.~Zhou, ``Generative hierarchical features from synthesizing images,'' in \emph{CVPR}, 2021.

\bibitem{alaluf2021restyle}
Y.~Alaluf, O.~Patashnik, and D.~Cohen-Or, ``{ReStyle}: A residual-based {StyleGAN} encoder via iterative refinement,'' in \emph{ICCV}, October 2021.

\bibitem{tov2021designing}
O.~Tov, Y.~Alaluf, Y.~Nitzan, O.~Patashnik, and D.~Cohen-Or, ``Designing an encoder for {StyleGAN} image manipulation,'' \emph{arXiv preprint arXiv:2102.02766}, 2021.

\bibitem{bdinvert}
S.~C. Kyoungkook~Kang, Seongtae~Kim, ``Bdinvert: {GAN} inversion for out-of-range images with geometric transformations.'' in \emph{ICCV}, 2021.

\bibitem{wang2021HFGI}
T.~Wang, Y.~Zhang, Y.~Fan, J.~Wang, and Q.~Chen, ``High-fidelity {GAN} inversion for image attribute editing,'' in \emph{CVPR}, 2022.

\bibitem{dinh2021hyperinverter}
T.~M. Dinh, A.~T. Tran, R.~Nguyen, and B.-S. Hua, ``{HyperInverter}: Improving {StyleGAN} inversion via hypernetwork,'' in \emph{CVPR}, 2022.

\bibitem{bai2022high}
Q.~Bai, Y.~Xu, J.~Zhu, W.~Xia, Y.~Yang, and Y.~Shen, ``High-fidelity {GAN} inversion with padding space,'' in \emph{ECCV}, 2022.

\bibitem{image2stylegan++}
R.~Abdal, Y.~Qin, and P.~Wonka, ``{Image2StyleGAN}++: How to edit the embedded images?'' \emph{arXiv preprint arXiv:1911.11544}, 2019.

\bibitem{gu2020image}
J.~Gu, Y.~Shen, and B.~Zhou, ``Image processing using multi-code {GAN} prior,'' in \emph{CVPR}, 2020.

\bibitem{pan2020exploiting}
X.~Pan, X.~Zhan, B.~Dai, D.~Lin, C.~C. Loy, and P.~Luo, ``Exploiting deep generative prior for versatile image restoration and manipulation,'' in \emph{ECCV}, 2020.

\bibitem{huh2020ganprojection}
M.~Huh, R.~Zhang, J.-Y. Zhu, S.~Paris, and A.~Hertzmann, ``Transforming and projecting images to class-conditional generative networks,'' in \emph{ECCV}, 2020.

\bibitem{roich2021pivotal}
D.~Roich, R.~Mokady, A.~H. Bermano, and D.~Cohen-Or, ``Pivotal tuning for latent-based editing of real images,'' \emph{ACM Trans. Graph.}, 2021.

\bibitem{pix2pix2017}
P.~Isola, J.-Y. Zhu, T.~Zhou, and A.~A. Efros, ``Image-to-image translation with conditional adversarial networks,'' \emph{CVPR}, 2017.

\bibitem{cgan}
M.~M. andvSimon Osindero, ``Conditional generative adversarial nets,'' \emph{CoRR}, 2014.

\bibitem{wang2018pix2pixHD}
T.-C. Wang, M.-Y. Liu, J.-Y. Zhu, A.~Tao, J.~Kautz, and B.~Catanzaro, ``High-resolution image synthesis and semantic manipulation with conditional gans,'' in \emph{CVPR}, 2018.

\bibitem{park2019SPADE}
T.~Park, M.-Y. Liu, T.-C. Wang, and J.-Y. Zhu, ``Semantic image synthesis with spatially-adaptive normalization,'' in \emph{CVPR}, 2019.

\bibitem{CycleGAN2017}
J.-Y. Zhu, T.~Park, P.~Isola, and A.~A. Efros, ``Unpaired image-to-image translation using cycle-consistent adversarial networks,'' in \emph{ICCV)}, 2017.

\bibitem{huang2018munit}
X.~Huang, M.-Y. Liu, S.~Belongie, and J.~Kautz, ``Multimodal unsupervised image-to-image translation,'' in \emph{ECCV}, 2018.

\bibitem{DRIT}
H.-Y. Lee, H.-Y. Tseng, J.-B. Huang, M.~K. Singh, and M.-H. Yang, ``Diverse image-to-image translation via disentangled representations,'' in \emph{ECCV}, 2018.

\bibitem{guidedI2I}
B.~AlBahar and J.-B. Huang, ``Guided image-to-image translation with bi-directional feature transformation,'' in \emph{ICCV)}, 2019.

\bibitem{DiscoGAN}
T.~Kim, M.~Cha, H.~Kim, J.~K. Lee, and J.~Kim, ``Learning to discover cross-domain relations with generative adversarial networks,'' in \emph{ICML}, 2017.

\bibitem{choi2018stargan}
Y.~Choi, M.~Choi, M.~Kim, J.-W. Ha, S.~Kim, and J.~Choo, ``Stargan: Unified generative adversarial networks for multi-domain image-to-image translation,'' in \emph{Proceedings of the IEEE Conference on Computer Vision and Pattern Recognition}, 2018.

\bibitem{yu2019multi}
X.~Yu, Y.~Chen, S.~Liu, T.~Li, and G.~Li, ``Multi-mapping image-to-image translation via learning disentanglement,'' in \emph{NeurIPS}, 2019.

\bibitem{choi2020starganv2}
Y.~Choi, Y.~Uh, J.~Yoo, and J.-W. Ha, ``Stargan v2: Diverse image synthesis for multiple domains,'' in \emph{CVPR}, 2020.

\bibitem{johnson2016perceptual}
J.~Johnson, A.~Alahi, and L.~Fei-Fei, ``Perceptual losses for real-time style transfer and super-resolution,'' in \emph{ECCV}, 2016.

\bibitem{vgg}
K.~Simonyan and A.~Zisserman, ``Very deep convolutional networks for large-scale image recognition,'' \emph{ICLR}, 2015.

\bibitem{wang2018esrgan}
X.~Wang, K.~Yu, S.~Wu, J.~Gu, Y.~Liu, C.~Dong, Y.~Qiao, and C.~Change~Loy, ``Esrgan: Enhanced super-resolution generative adversarial networks,'' in \emph{Proceedings of the European conference on computer vision (ECCV) workshops}, 2018.

\bibitem{ledig2017photo}
C.~Ledig, L.~Theis, F.~Husz{\'a}r, J.~Caballero, A.~Cunningham, A.~Acosta, A.~Aitken, A.~Tejani, J.~Totz, Z.~Wang \emph{et~al.}, ``Photo-realistic single image super-resolution using a generative adversarial network,'' in \emph{CVPR}, 2017.

\bibitem{yu2015lsun}
F.~Yu, A.~Seff, Y.~Zhang, S.~Song, T.~Funkhouser, and J.~Xiao, ``{LSUN}: Construction of a large-scale image dataset using deep learning with humans in the loop,'' \emph{arXiv preprint arXiv:1506.03365}, 2015.

\bibitem{resnet}
K.~He, X.~Zhang, S.~Ren, and J.~Sun, ``Deep residual learning for image recognition,'' in \emph{CVPR}, 2016.

\bibitem{howard2017mobilenets}
A.~G. Howard, M.~Zhu, B.~Chen, D.~Kalenichenko, W.~Wang, T.~Weyand, M.~Andreetto, and H.~Adam, ``Mobilenets: Efficient convolutional neural networks for mobile vision applications,'' \emph{arXiv preprint arXiv:1704.04861}, 2017.

\bibitem{adain}
X.~Huang and S.~Belongie, ``Arbitrary style transfer in real-time with adaptive instance normalization,'' in \emph{ICCV}, 2017.

\bibitem{bartz2020one}
C.~Bartz, J.~Bethge, H.~Yang, and C.~Meinel, ``One model to reconstruct them all: A novel way to use the stochastic noise in {StyleGAN},'' \emph{arXiv preprint arXiv:2010.11113}, 2020.

\bibitem{unet}
O.~Ronneberger, P.~Fischer, and T.~Brox, ``U-net: Convolutional networks for biomedical image segmentation,'' in \emph{International Conference on Medical image computing and computer-assisted intervention}.\hskip 1em plus 0.5em minus 0.4em\relax Springer, 2015, pp. 234--241.

\bibitem{zhu2022resefa}
J.~Zhu, Y.~Shen, Y.~Xu, D.~Zhao, and Q.~Chen, ``Region-based semantic factorization in {GAN}s,'' in \emph{International Conference on Machine Learning (ICML)}, 2022.

\end{thebibliography}
\begin{IEEEbiography}[{\includegraphics[width=1in,height=1.25in,clip,keepaspectratio]{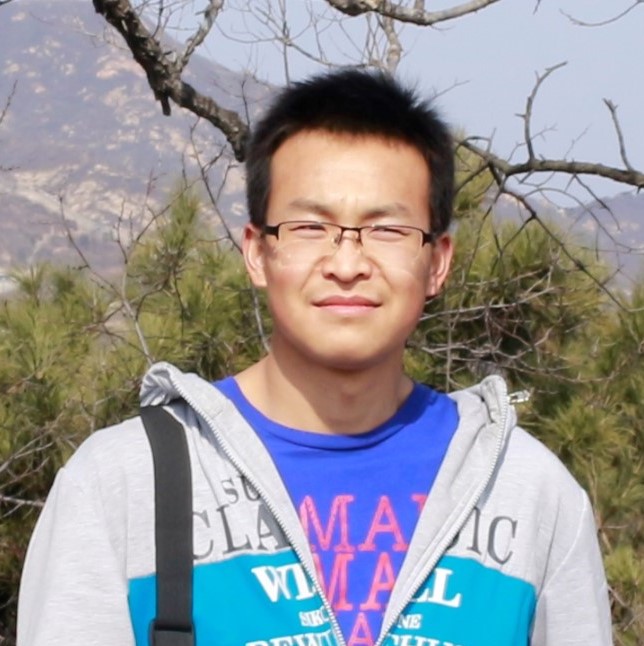}}]{Jiapeng Zhu}
is a first-year Ph.D. student at the Department of Computer Science and Engineering at the Hong Kong University of Science and Technology. His research interests include computer vision and generative models.
\end{IEEEbiography}

\begin{IEEEbiography}[{\includegraphics[width=1in,height=1.25in,clip,keepaspectratio]{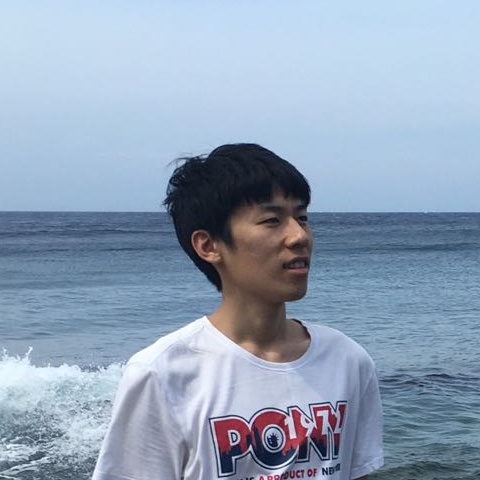}}]{Yujun Shen}
is a senior researcher at Ant Research. Before that, he worked as a senior researcher at ByteDance Inc. He received his Ph.D. degree at the Chinese University of Hong Kong and his B.S. degree at Tsinghua University. His research interests include computer vision and deep learning, particularly in 3D vision and generative models. He is an award recipient of the Hong Kong Ph.D. Fellowship.
\end{IEEEbiography}

\begin{IEEEbiography}[{\includegraphics[width=1in,height=1.25in,clip,keepaspectratio]{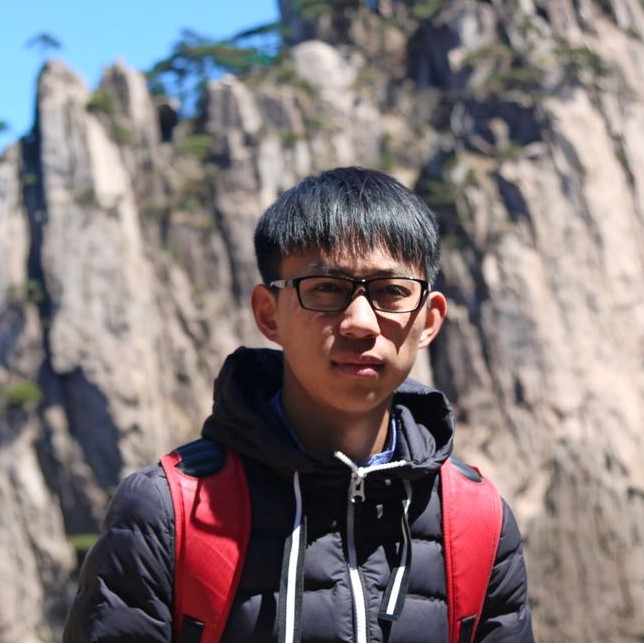}}]{Yinghao Xu}
is a fourth-year Ph.D. student at Multimedia Lab (MMLab), Department of Information Engineering in The Chinese University of Hong Kong. He graduated from Zhejiang University in 2019. His research interests include video understanding, generative models, and structural representation for vision perception.
\end{IEEEbiography}

\begin{IEEEbiography}[{\includegraphics[width=1in,height=1.25in,clip,keepaspectratio]{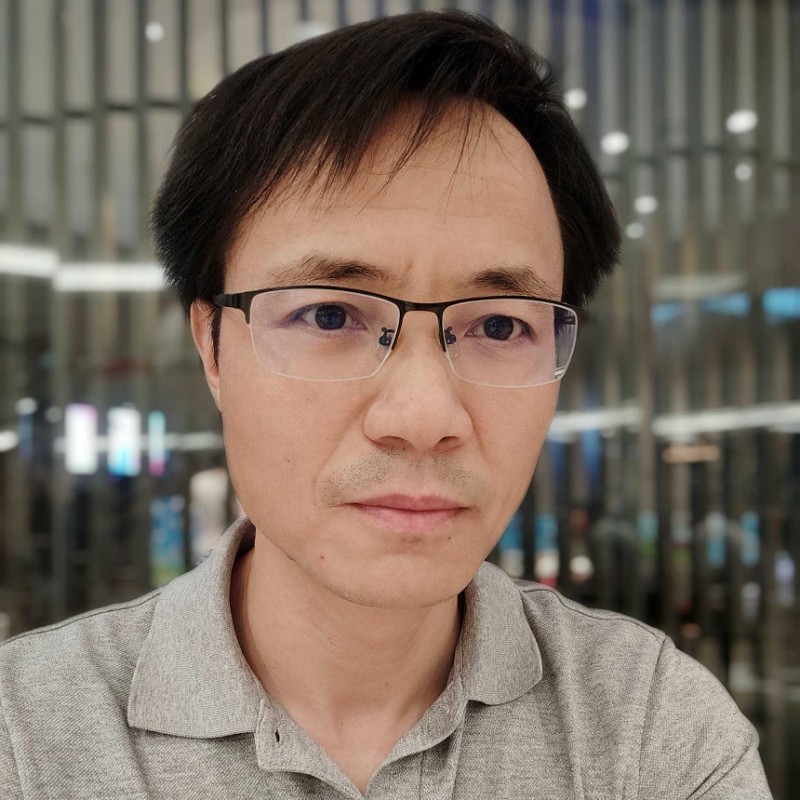}}]{Deli Zhao}
is a director leading the Visual Intelligence Lab at Ant Research. He graduated from Shanghai Jiao Tong University in 2006. He ever worked at the Visual Computing Group in Microsoft Research Asia and  Multimedia Lab (MMLab) in the Chinese University of Hong Kong.
His research interests are mainly focused on generative models, multimodal learning, foundation models, and visual intelligence.
\end{IEEEbiography}

\begin{IEEEbiography}[{\includegraphics[width=1in,height=1.25in,clip,keepaspectratio]{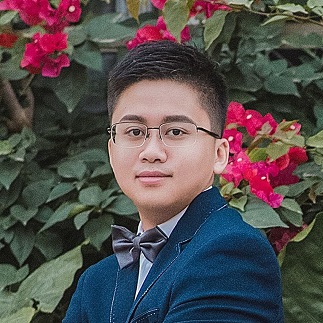}}]{Qifeng Chen} is an assistant professor of the Department of Computer Science and Engineering and the Department of Electronic and Computer Engineering at The Hong Kong University of Science and Technology. He received his Ph.D. in computer science from Stanford University in 2017. His research interests include image processing and synthesis and 3D vision. He won the MIT Tech Review's 35 Innovators under 35 in China and the Google Faculty Research Award in 2018. 
\end{IEEEbiography}

\begin{IEEEbiography}[{\includegraphics[width=1in,height=1.25in,clip,keepaspectratio]{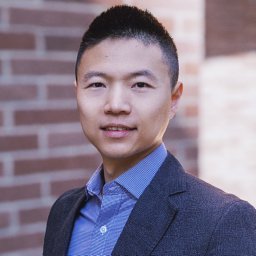}}]{Bolei Zhou}
is an assistant professor in the Computer Science Department at the University of California, Los Angeles (UCLA). He received his Ph.D. in computer science at the Massachusetts Institute of Technology. His research is on machine perception and autonomy. He received the MIT Tech Review’s Innovators under 35 in Asia-Pacific Award and Intel Rising Star Faculty Award, and his research was featured in media outlets such as TechCrunch, Quartz, and MIT News.
\end{IEEEbiography}
\newpage


\section*{A1. Semantic Diffusion Analysis.} \label{sec:supp-semantic-diffusion}
\begin{figure}[th]
  \centering
  \includegraphics[width=1.0\linewidth]{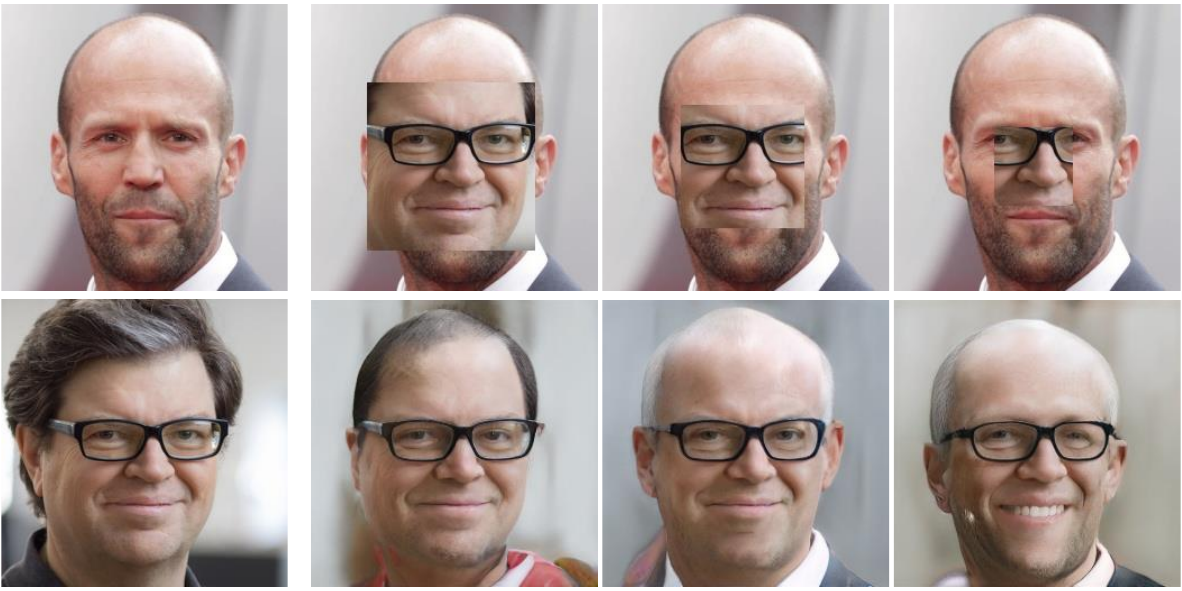}
  \vspace{-16pt}
  \caption{
    The effect of crop size on semantic diffusion.
    Top-left corner shows the context image, while the bottom-left corner shows the target image.
    Each remaining column corresponds to a different crop size.
    The top row shows the direct copy-paste results, while the bottom row shows the semantic diffusion results.
  }
  \label{fig:face-diffusion-size}
\end{figure}

We further studied the impact of crop size on semantic diffusion.
As shown in \cref{fig:face-diffusion-size}, we can see that the larger the crop size is (\emph{i.e.}, the larger reference region from the target face), the better the identity information is preserved.
For example, on the second column of \cref{fig:face-diffusion-size}, even hair is transmitted from the target image to the context image since the temples are included in the cropped patch.
In the last column, however, the diffused result is no longer like the target face at all (\emph{e.g.}, the facial shape and mouth).
That is because, during the process of masked optimization, only the foreground patch is used as a reference.
The surroundings will adaptively change starting from the encoder initialization.
Even so, thanks to the \emph{in-domain} property, our approach is still able to complete the entire eyeglasses and generate a smooth diffusion result.

\section*{A2.Visualizing the Initial Inversion Space.} \label{sec:vis-initial-inversion-space}

\begin{figure}[th]
  \centering
  \includegraphics[width=0.98\linewidth]{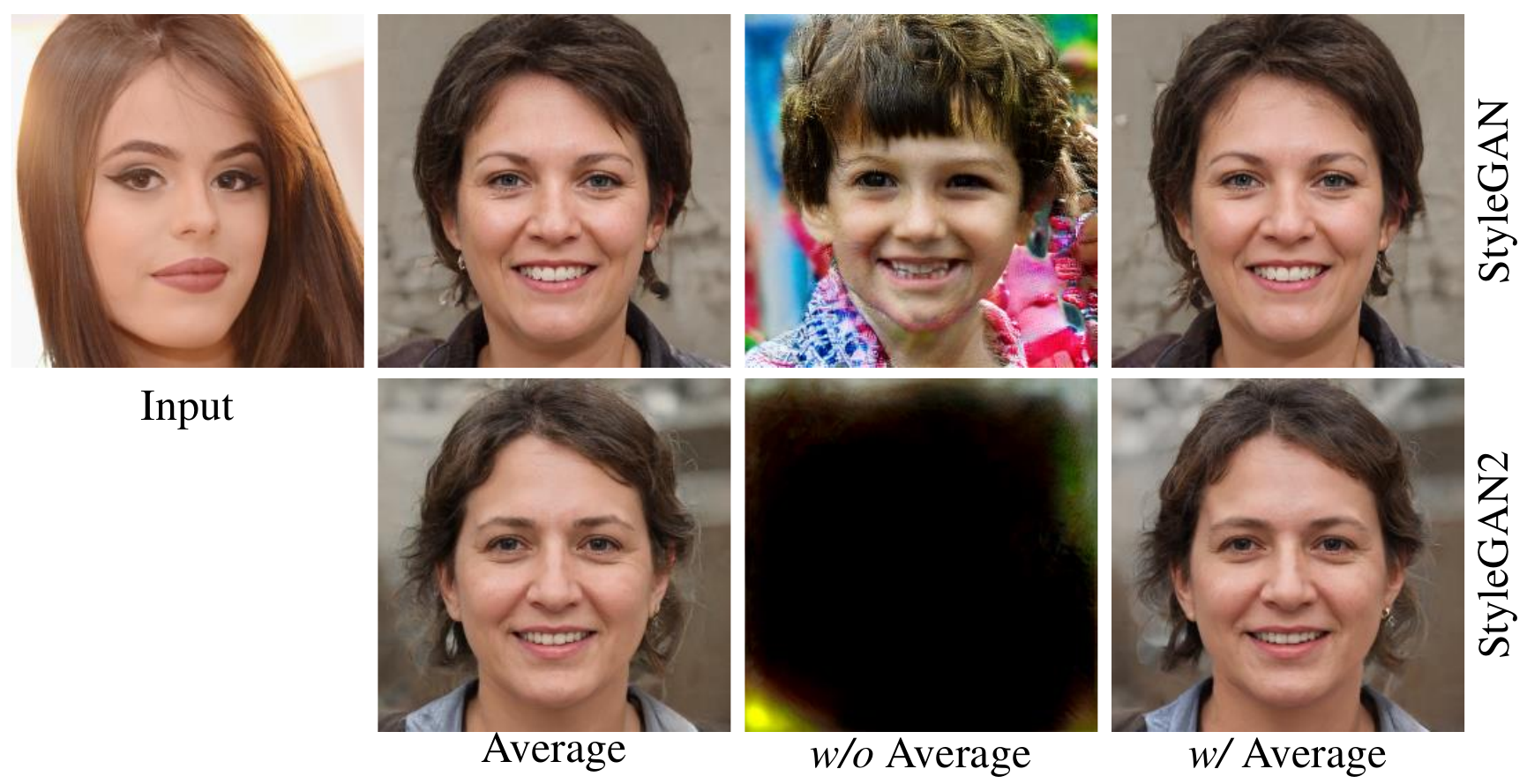}
  \vspace{-10pt}
  \caption{
    \ReviseSecond{The initial reconstruction on the StyleGAN~\cite{stylegan} and StyleGAN2~\cite{stylegan2} whether or not to use average value during training.
    The images above the average mean the synthesized images concerning the average latent code on each model.
    }
  }
  \label{fig:initial-rec}
\end{figure}

\cref{fig:initial-rec} shows the impact of adding the average latent code $\bar{\bf{w}}$ when training the encoder for StyleGAN~\cite{stylegan} and StyleGAN2~\cite{stylegan2}.
We observe that, in StyleGAN~\cite{stylegan}, even without the average latent code, the initial reconstruction is a human face, indicating that the initial latent code produced by the encoder lies in or near the native latent distribution.
When the average latent code is added, the initial latent code is shifted to another latent code that is very close to the $ \bar{\tb{w}} $, which of course, lies in the native latent distribution.
In both cases, the encoder is easy to learn.
In contrast, on StyleGAN2, the initial reconstruction is a non-human face or even an abnormal image, suggesting that the initial latent code may be far away from the native latent distribution and thereby difficulting the learning process.
When the average latent code is added, the initial latent code is dominated by $\bar{\bf{w}}$ and pulled back toward the native latent space, which eases the process of training.

\begin{figure}[th]
  \centering
  \includegraphics[width=1.0\linewidth]{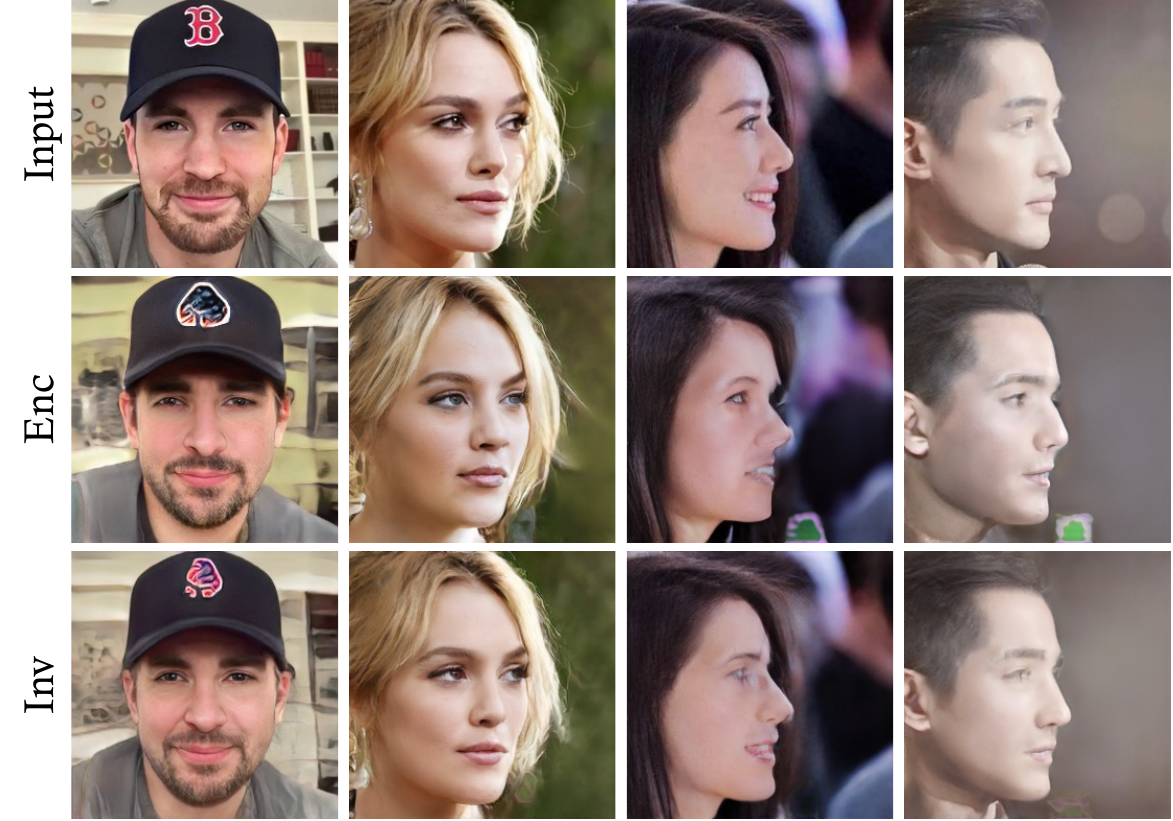}
  \vspace{-15pt}
  \caption{
    Extreme samples inversion using our method.
    The first row shows the input images, while the second and third rows show the results from our encoder and in-domain inversion, respectively.
  }
  \label{fig:failure-case}
\end{figure}

\section*{A3. Experiments on the Additional Datasets.} \label{sec:additional-res}

\begin{figure*}[!ht]
  \centering
  \includegraphics[width=1.0\linewidth]{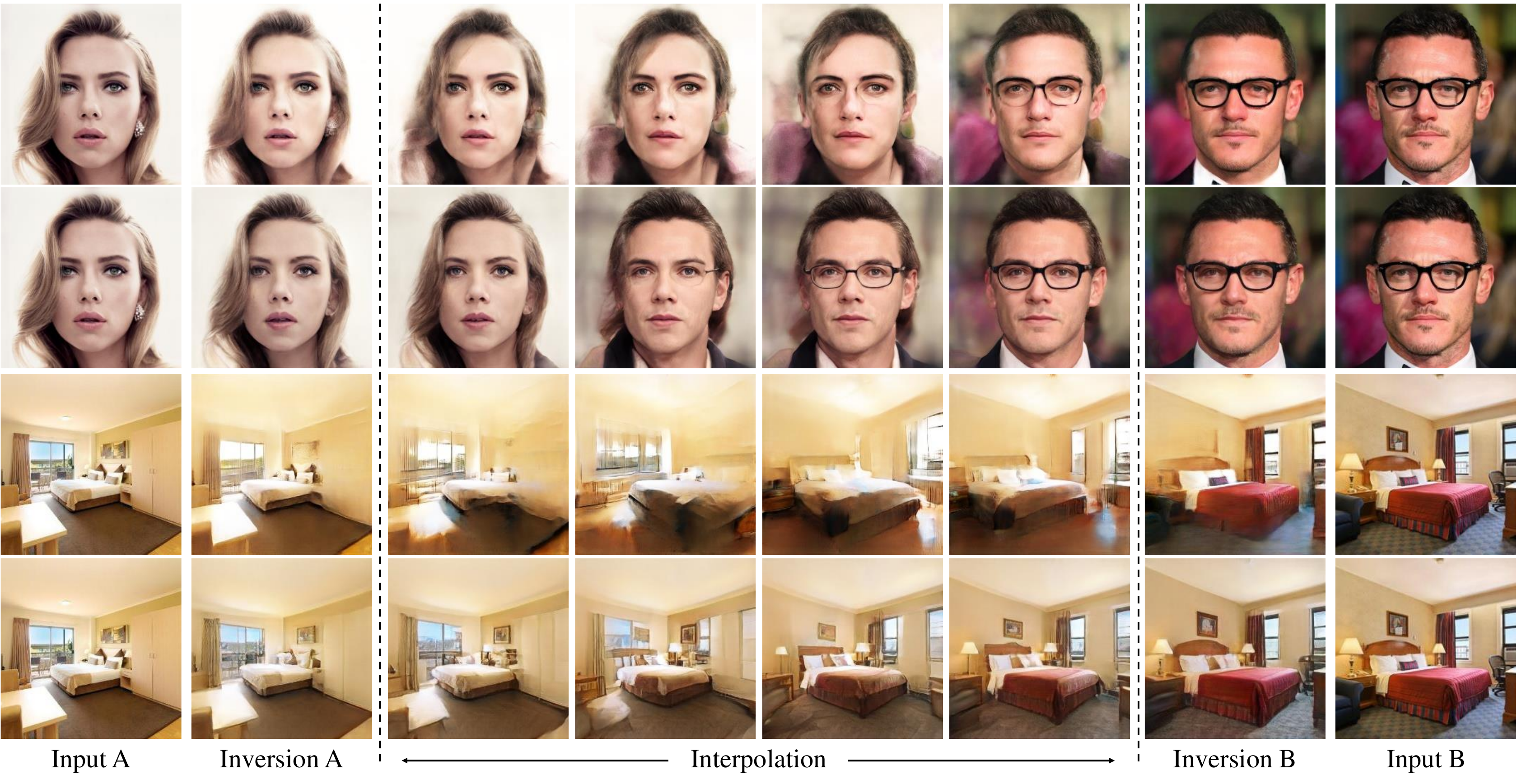}
  \vspace{-22pt}
  \caption{
    Qualitative comparison on image interpolation between Image2StyleGAN \cite{image2stylegan} (odd rows) and our \emph{in-domain} inversion (even rows).
  }
  \label{fig:interpolation-ffhq-bedroom}
  \vspace{-12pt}
\end{figure*}

\noindent\textbf{Image Interpolation.} In addition to the tower interpolation results presented in the main text, we have included additional interpolation results for the face and bedroom datasets, as shown in \cref{fig:interpolation-ffhq-bedroom}.
For the face dataset, our method achieves much smoother interpolated faces than Image2StyleGAN.
For example, in the first two rows of \cref{fig:interpolation-ffhq-bedroom}, eyeglasses are distorted during the interpolation process with Image2StyleGAN, and the change from female to male is unnatural.
For bedroom images, the interpolation results from Image2StyleGAN exhibit artifacts and blurriness.
Instead, our inverted codes lead to more satisfying interpolation.

\begin{figure*}[th]
  \centering
  \includegraphics[width=1\linewidth]{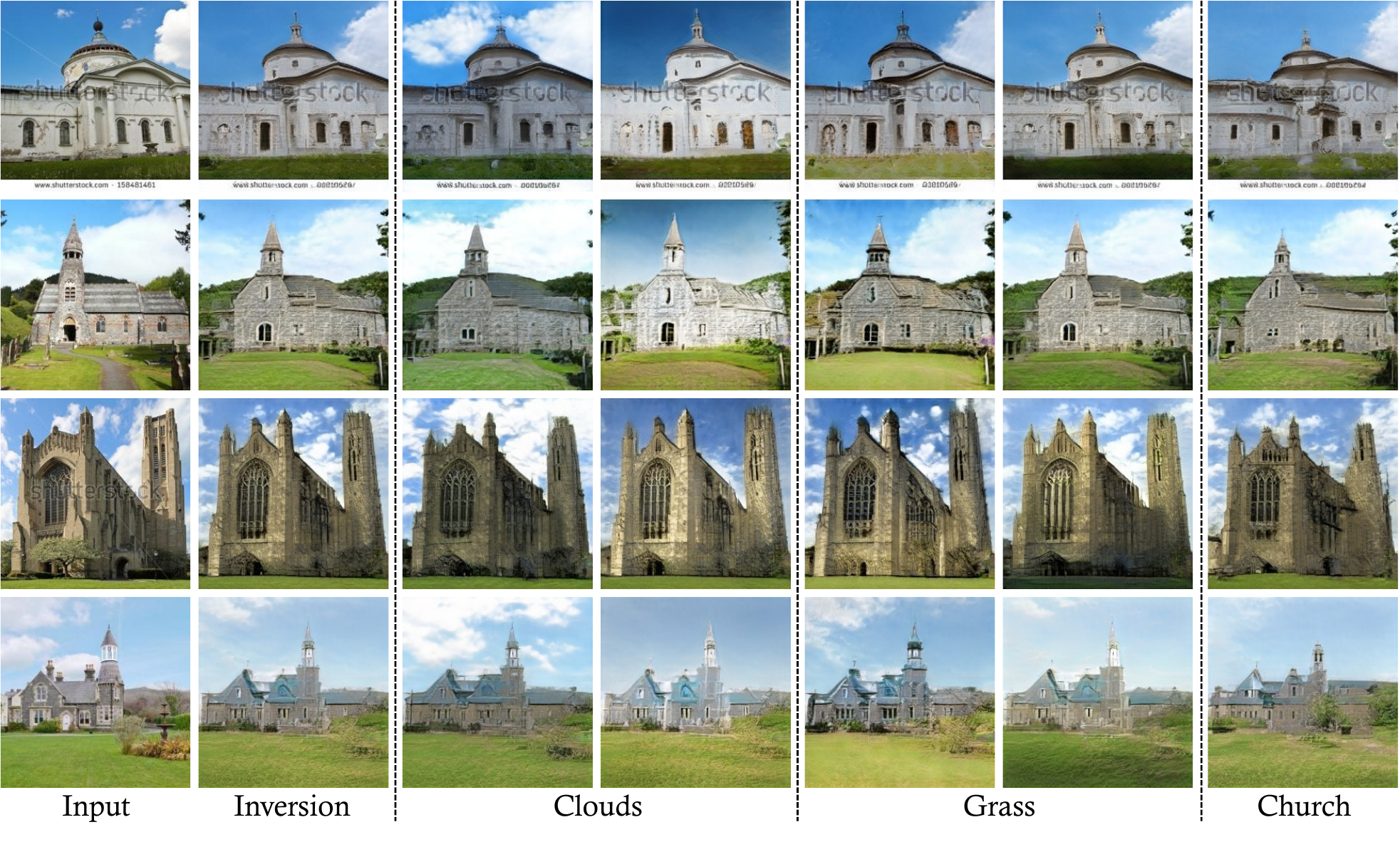}
  \vspace{-20pt}
  \caption{
    Inversion and editing results on the LSUN church dataset.
    The first column shows the input images, the second column shows the results from our encoder, and the rest results are the editing results using the method proposed in \cite{zhu2022resefa}.
  }
  \label{fig:church-inversion}
\end{figure*}

\noindent\textbf{Semantic Manipulation.} In the main text, we have performed experiments to assess the effectiveness of our proposed method using various datasets, including human faces, indoor scenes (\textit{i.e.}, bedrooms), and outdoor scenes (\textit{i.e.}, towers). 
In addition to these datasets, we also include results from the LSUN churches dataset to support our findings further. 
The results obtained from this additional dataset, as shown in \cref{fig:church-inversion}, align with the consistent conclusions drawn from the datasets discussed in the main text.

\section*{A4. Failure Case.} \label{sec:failure-case}

Since our assumption is that only good editability can be achieved when the inverted code lands inside the original domain, the proposed method has the limitation that only the images with the same distribution as the training set can be well inverted and manipulated.
For those images with major differences from the training set, it is hard to invert them well, let alone edit them.
\cref{fig:failure-case} shows some results using our method, which shows that the reconstruction performance is not good when the given images are far away from the training set.
For example, the letter on the man's hat and the earrings on the woman can not be accurately reconstructed.
Also, the reconstruction results are poor for the faces with a large pose, as the last two columns show in \cref{fig:failure-case}.

\setlength{\tabcolsep}{5.0pt}
\begin{table*}[th]
  \setlength{\tabcolsep}{5.5pt}
  \caption{
          The quantitative results on StyleGAN \cite{stylegan} and StyleGAN2 \cite{stylegan2} on LSUN tower and bedroom datasets with generator retrained.
  }
  \label{tab:val-on-bedroom-tower}
  \vspace{-8pt}
  \centering\small
  \begin{tabular}{l|cccc|cc|cccc|cc}
    \toprule
      &  \multicolumn{6}{c|}{StyleGAN \cite{stylegan}}
      &  \multicolumn{6}{c}{StyleGAN2 \cite{stylegan2}} \\
    \midrule
      &  \multicolumn{4}{c|}{Reconstruction}
      &  \multicolumn{2}{c|}{Interpolation}
      &  \multicolumn{4}{c|}{Reconstruction}
      &  \multicolumn{2}{c}{Interpolation}              \\
    \midrule
 \diagbox[width=8em]{Dataset}{Metrics}
        &    \FID    &    \SWD    &    \MSE    &   \SSIM    &    \FID    &   \SWD
        &    \FID    &    \SWD    &    \MSE    &   \SSIM    &    \FID    &   \SWD       \\
      \midrule
 Tower
        &    12.41   &   19.50    &    0.093   &    0.500   &   33.02    &  29.74
        &    8.55    &   18.60    &    0.091   &    0.504   &   32.19    &  40.80       \\
      \midrule
 Bedroom
        &  9.61     &    21.93   &    0.079    &    0.521    & 19.63     & 21.40
        &  6.56     &    19.73   &    0.078    &    0.528    & 20.13     & 24.20        \\
      \bottomrule
  \end{tabular}
  \vspace{-5pt}
\end{table*}

\end{document}